\documentclass[sn-mathphys-ay]{sn-jnl}

\usepackage{graphicx}
\usepackage{multirow}
\usepackage{amsmath, amssymb, amsfonts}
\usepackage{amsthm}
\usepackage{mathrsfs}
\usepackage[title]{appendix}
\usepackage{xcolor}%
\usepackage{textcomp}%
\usepackage{manyfoot}%
\usepackage{booktabs}%
\usepackage{algorithm}
\usepackage{algorithmic}
\usepackage{listings}%
\usepackage{array}
\usepackage{bm}
\usepackage{subfigure}
\usepackage{hyperref}

\newcommand{\ba}{\mathbf{a}}
\newcommand{\bb}{\mathbf{b}}
\newcommand{\bc}{\mathbf{c}}
\newcommand{\bd}{\mathbf{d}}

\newcommand{\cX}{\mathbf{X}}
\newcommand{\cY}{\mathbf{Y}}

\newcommand{\cE}{\mathbf{E}}
\newcommand{\cA}{\mathbf{A}}
\newcommand{\cB}{\mathbf{B}}
\newcommand{\cC}{\mathbf{C}}
\newcommand{\cD}{\mathbf{D}}

\newcommand{\eX}{\bm{\mathscr{X}}}
\newcommand{\eY}{\bm{\mathscr{Y}}}
\newcommand{\eS}{\bm{\mathscr{S}}}

\newtheorem{theorem}{Theorem}

\newtheorem{prop}{Proposition}

\numberwithin{equation}{section}
\title{Multiple Linked Tensor Factorization}
\author[1]{\fnm{Zhiyu} \sur{Kang}}
\author[2]{\fnm{Raghavendra B. } \sur{Rao}}
\author*[1]{\fnm{Eric F.} \sur{Lock}}\email{elock@umn.edu}

\affil[1]{\orgdiv{Division of Biostatistics and Health Data Science}, \orgname{School of Public Health, University of Minnesota}, \orgaddress{\city{Minneapolis}, \postcode{55455}, \state{MN}, \country{USA}}}
\affil[2]{\orgdiv{Division of Neonatology}, \orgname{Department of Pediatrics, University of Minnesota}, \orgaddress{\city{Minneapolis}, \postcode{55455}, \state{MN}, \country{USA}}}

\begin{document}

\abstract{In biomedical research and other fields, it is now common to generate high content data that are both \emph{multi-source} and \emph{multi-way}.  Multi-source data are collected from different high-throughput technologies while multi-way data are collected over multiple dimensions, yielding multiple tensor arrays.  Integrative analysis of these data sets is needed, e.g., to capture and synthesize different facets of complex biological systems. However, despite growing interest in multi-source and multi-way factorization techniques, methods that can handle data that are both multi-source and multi-way are limited.  In this work, we propose a Multiple Linked Tensors Factorization (MULTIFAC) method extending the CANDECOMP/PARAFAC (CP) decomposition to simultaneously reduce the dimension of multiple multi-way arrays and approximate underlying signal. We first introduce a version of the CP factorization with $L_2$ penalties on the latent factors, leading to rank sparsity.  When extended to multiple linked tensors, the method  automatically reveals latent components that are shared across data sources or  individual to each data source. We also extend the decomposition algorithm to its expectation–maximization (EM) version to handle incomplete data with imputation. Extensive simulation studies are conducted to demonstrate MULTIFAC’s ability to (i) approximate underlying signal, (ii) identify shared and unshared structures, and (iii) impute missing data.  The approach yields an interpretable decomposition on multi-way multi-omics data for a study on early-life iron deficiency. }


\maketitle

\section{Introduction}
In many scientific disciplines, the representation and analysis of high-content data with complex structure has become critical.  Tensors, which extend traditional two-way matrices to multi-dimensional arrays, are often well-suited to capture the rich content inherent in such data. For example, as a motivating application we consider a study on early life iron deficiency in a cohort of infant rhesus monkeys \citep{rao_lock_etal_2023}. Hematology data are collected at multiple developmental time points from a group of monkey, and the resulting data are best represented by a three-way tensor data: $Monkeys \times Age\times Hematology \; Indices$. 

Similar to how a matrix decomposition captures latent structures within two-dimensional data, tensor decomposition techniques can be used to uncover the underlying signals from noise, thereby reducing the dimensionality and enhancing interpretability of multi-way data arrays. One traditional and widely used method is the CANDECOMP/PARAFAC (CP) decomposition \citep{harshman_1970, carroll_chang_1970}. The CP decomposition represents a tensor as a sum of multiple outer products of vectors, each corresponding to a dimension of the tensor. This method can be viewed as an extension of matrix decomposition, such as singular value decomposition (SVD) and principal component analysis (PCA), to a higher-order version that enables analysis of additional ways such as time, tissue types or other dimensions in the dataset. 

Often, multiple high dimensional datasets are collected for a single study that capture different facets of the research subjects. This motivates a large number of integrative data analysis methods that combine multiple datasets linked in some ways. Matrix data can be vertically or horizontally linked, sharing the same features or samples across datasets. A simple example for linked data that share the same samples is multi-omics data,
in which data for multiple molecular platforms (protoeomics, metabolomics, genomics, etc.) are measured for the same sample set.
Under this scenario, there are several methods that decompose each omics dataset into a sum of shared structures and individual structures \citep{lock_jive_2013, gaynanova_li_slide_2019, sjive_2022,argelaguet2018multi}, where the shared structure is represented by a latent factorization (analogous to PCA) with the same score matrix across datasets, and individual structure is defined by each dataset's unique loading and score matrices. Other methods take a similar approach to decompose data with the same features measured across different sample sets \citep{wang_lock_2024,de2021bayesian}.  
Bidimensional Integrative Factorization (BIDIFAC) further extends this framework by simultaneously factorizing bidimensionally linked matrices, enabling the capture of structures shared both vertically and horizontally \citep{park_lock_2019, lock_park_2020}. 

Beyond matrices, there is a small but growing literature on the integrative analysis of multiple data with tensor structure. These scenarios often involve collecting multiple tensor datasets, some of which share one or more tensor dimensions. The coupled matrix and tensor decomposition (CMTF) method is a special case for the integration of a matrix and a tensor, which offers flexibility in handling heterogeneous data but lacks discussion on the choice of tensor ranks \citep{acar_2011, acar_2013}. Further advancements in the field are represented by structured data fusion (SDF) and Bayesian multi-tensor decomposition \citep{sorber_sdf_2015, khan_bayesian_2014}. The SDF approach describes a very general framework to joint fusing factorization of multiple matrices or tensors. The Bayesian multi-tensor factorization method provides a robust framework for decomposing multiple matrices and tensors into shared and individual factors by the spike and slab prior. However, these methods  do not focus on missing data imputation.  Often, tensors are incomplete, and accommodating missing data and imputing missing values are critical for a factorization approach. In the context of multiple linked tensors, a commonly encountered special case is tensor-wise missing, in which data are missing for an tensor for some samples. 

For our motivating application to early life iron deficiency, there is also diffusion tensor imaging (DTI) data available for the monkeys at a single timepoint over multiple brain regions. Thus the data have the form of two tensors, $\textit{Monkeys} \times \textit{Age} \times \times{Hematology Indices}$ and $\textit{Monkeys} \times \textit{Brain Regions}  \times \textit{DTI Parameters}$, that are linked across the same monkeys.  The data have complex missing structure, as the hematology data are incomplete or not collected for all timepoints for some monkeys, and not all monkeys have both tensors (hematology and DTI) observed.   

We propose a novel approach to tensor and multiple tensor factorization, leveraging new results on penalized CP factorization. Specifically, we show that an $L_2$ penalty on the individual factor matrices for a CP factorization is equivalent to an $L_p$ ($p<1$) penalty on the singular values of the tensor approximation, promoting rank sparsity. The $L_2$ penalty is computationally straightforward to optimize, and effectively identifying the tensor ranks and shared and individual signal.   
   
The motivation behind our proposed MULTIFAC method stems from the need to address specific challenges in iron deficiency and other real-world data applications. The purpose is several fold: one goal is dimension reduction, aiming to find an efficient way to represent multiple tensors while uncover the underlying signal. Secondly, compared with applying single factorization to each tensor, our method can reveal shared structure across the data for interpretation. Last but not least, our method can properly address both tensor-wise missing and entry-wise missing problems, providing a flexible and powerful tools for imputation.

\section{Notation and Preliminaries}
The order, ways or modes of a tensor is the number of dimensions. Throughout this paper, scalars are denoted by lowercase letters ($a$), vectors (one-way tensor) are denoted by boldface lowercase letters (${\mathbf{a}}$), matrices (two-way tensor) are denoted by boldface capital letters ($\mathbf{A}$), and high-order tensors are denoted by Euler script letters ($\bm{\mathscr{X}}$).  For an $N$-way tensor $\eX \in \mathbb{R}^{I_1\times \cdots \times I_N}$, its entry ($i_1, \ldots, i_N$) is denoted by $x_{i_1\ldots i_N}$.

Given two matrices $\cA \in \mathbb{R}^{I_1\times I_3}$ and $\cB \in \mathbb{R}^{I_2\times I_3}$, the Khatri-Rao product of them, denoted as $\cA \odot \cB$, is defined as their “matching columnwise” Kronecker product. The result is a matrix of size $(I_1I_2)\times I_3$ defined by
$$
\mathbf{A} \odot \mathbf{B} = \left[ \mathbf{a}_1 \otimes \mathbf{b}_1 \quad \mathbf{a}_2 \otimes \mathbf{b}_2 \quad \cdots \quad \mathbf{a}_{I_3} \otimes \mathbf{b}_{I_3} \right],
$$
where $\otimes$ denotes the Kronecker product. The Kronecker product of matrices $\cA\in \mathbb{R}^{I_1\times I_2}$ and $\cB\in \mathbb{R}^{I_3\times I_4}$ is a matrix of size $(I_1I_3) \times (I_2I_4)$ defined by

\begin{align*}
    \cA \otimes \cB &=
\begin{bmatrix}
    a_{11} \cB & a_{12} \cB & \cdots & a_{1I_2} \cB \\
    a_{21} \cB & a_{22} \cB & \cdots & a_{2I_2} \cB \\
    \vdots & \vdots & \ddots & \vdots \\
    a_{I_11} \cB & a_{I_12} \cB & \cdots & a_{I_1I_2} \cB
\end{bmatrix}\\
&=
\begin{bmatrix}
    \ba_1 \otimes \bb_1 & \ba_1 \otimes \bb_2 & \ba_1 \otimes \bb_3 & \cdots & \ba_{I_2} \otimes \bb_{I_4-1} & \ba_{I_2} \otimes \bb_{I_4}
\end{bmatrix}.
\end{align*}

The Frobenius norm of a tensor $\eX$ is defined by the square root of the sum of the squares of all the elements
$$
\| \eX \|_F = \sqrt{\sum^{I_1}_{i_1=1} \cdots \sum^{I_N}_{i_N=1} x_{i_1\ldots i_N}^2}.
$$

Given vectors $\ba_1 = [a_{11}, a_{12}, \ldots, a_{1I_1}]^\top, \ba_2 = [a_{21}, a_{22}, \ldots, a_{2I_2}]^\top, \ldots, \mathbf{a}_N = [a_{N1}, a_{N2}, \ldots, a_{NI_N}]^\top$ of length $I_1, \ldots, I_N$ respectively, the outer product
$$
\bm{\mathscr{X}} = \ba_1 \circ \ba_2 \circ \cdots \circ  \ba_N
$$
defines a $N$-way rank-1 tensor of dimensions $I_1\times \cdots \times I_N$ with entries
$$
x_{i_1, \ldots, i_N} = a_{1i_1} a_{2i_2}\cdots a_{Ni_N}, 1\leq i_n \leq I_N.
$$

Given two tensors $\eX$ and $\eY$ of the same dimensions $I_1\times \cdots \times I_N$, the Hadamard product is denoted by $\eX * \eY \in \mathbb{R}^{I_1\times \cdots \times I_N}$ and defined as 
$$
(\eX * \eY)_{i_1i_2\ldots i_N} = x_{i_1i_2\ldots i_N} y_{{i_1i_2\ldots i_N}}.
$$

The $n$-mode matricization or unfolding of tensor $\eX$, denoted by $\cX_{(n)}$, rearranges $\eX$ as a matrix by using the mode-$n$ ``fibers" as the columns of the resulting matrix (see \citet{kolda_tensor_review_2009} for details).

\subsection{Matrix Decomposition and Results}
Here we present some well-established results for a matrix $\cX \in \mathbb{R}^{I_1 \times I_2}$, to lay the groundwork for novel extensions of these results for a tensor. Define the SVD for $\cX$  as:
$$
\cX = \tilde\cA_1 \cD \tilde\cA_2^\top,
$$
where $\tilde \cA_1 \in \mathbb{R}^{I_1 \times R}$ and $\tilde\cA_2 \in \mathbb{R}^{I_2 \times R}$ have orthonormal columns, $\cD \in \mathbb{R}^{R \times R}$ is a diagonal matrix containing the singular values, and $R = \min(I_1, I_2)$ is the rank of matrix. With SVD, one can obtain a low-rank approximation of a matrix $\cX$ by only keeping the largest $k$ singular values and corresponding columns in $\tilde\cA_1$ and $\tilde\cA_2$:
$$
\hat{\cX} = \sum_{r=1}^{k} d_r \tilde\ba_{1r} \tilde\ba_{2r}^\top,
$$
where $d_r$ is the $r$-th largest singular value, and $\tilde\ba_{1r}$ and $\tilde\ba_{2r}$ are the corresponding left and right singular vectors.

The nuclear norm $\|\cX\|_*$ of a matrix $\cX$ is defined as the sum of its singular values:
$$
\|\cX\|_* = \sum_{r=1}^{R} d_r.
$$
Minimizing the nuclear norm can be used to promote low-rank solutions in matrix approximation problems. A popular approach is to penalize squared error loss with the nuclear norm, leading to the following optimization problem:
\begin{equation} \label{svd_nuclear}
    \min_{\hat{\cX}} \frac{1}{2} \|\cX - \hat{\cX}\|_F^2 + \sigma \|\hat{\cX}\|_*.
\end{equation}
The solution to the above optimization problem is given by the soft-thresholding operator on the singular values of $\cX$, as in Proposition \ref{prop:1} \citep{mazumder_2010}. 
\begin{prop} \label{prop:1}
    Let $\cX = \tilde\cA_1 \cD \tilde\cA_2^{\top}$ be the SVD of $\cX$. Then, the optimal solution to problem \eqref{svd_nuclear} $\hat{\cX}$ is given by
\begin{equation}
    \hat\cX = \tilde\cA_1 \hat \cD \tilde\cA_2^{\top},
\end{equation}
where the diagonal elements of $\hat\cD$ are $\hat d_r = \max(d_r - \sigma, 0)$, $r = 1, \ldots, R$.
\end{prop}
By choosing a proper penalty factor $\sigma$, this soft-thresholding property results in a low-rank approximation of $\cX$. We can further rewrite the matrix decomposition as $\cX = \cA_1 \cA_2^{\top}$, where the singular value matrix $\cD$ is absorbed into the factor matrices $\cA_1$ and $\cA_2$. The equivalence of nuclear norm penalization and Frobenius norm ($L_2$) penalization applied on $ \cA_1 $ and $\cA_2$ is established in Proposition \ref{prop:2} \citep{mazumder_2010}.
\begin{prop}\label{prop:2}
    For a matrix $\cX \in \mathbb{R}^{I_1 \times I_2}$, we have
    \begin{equation}\label{nuclear_frob}
    \min_{\hat{\cX}} \|\cX - \hat{\cX}\|_F^2 + 2\sigma \| \hat{\cX}\|_* = \min_{\cA_1:I_1\times r, \cA_1:I_2\times r} \|\cX - \cA_1\cA_2^\top\|_F^2 + \sigma  \left( \|\cA_1\|_F^2 + \|\cA_2\|_F^2 \right),
\end{equation}
where $r = \text{min}(I_1, I_2)$, $ \hat{\cX} = \hat \cA_1\hat \cA_2^\top$ solves the left-hand size of \ref{nuclear_frob} and $\{\hat \cA_1, \hat \cA_2\}$ solves the right-hand size of \ref{nuclear_frob}.
\end{prop}

Proposition \ref{prop:2} shows the ability of a $L_2$ penalty on the factor matrices to find a low-rank representation, which motivates our penalty for tensor decompositions.

\subsection{Single Tensor decomposition}
One of the most commonly used models for tensor decomposition is the CANDECOMP/PARAFAC (CP) decomposition \citep{carroll_chang_1970, harshman_1970}. Analogous to expressing a matrix as a sum of outer products of two vectors, CP decomposition expresses a tensor as a sum of outer product of multiple vectors, i.e., rank-one tensors.  Given a $N$-order tensor $\eX \in \mathbb{R}^{I_1\times \cdots \times I_N}$, $\eX$ may be approximated with $\hat \eX$ that has a rank-$R$ CP decomposition as follows:
\begin{equation}
    \eX \approx \sum_{r=1}^{R} \ba_{1r} \circ \ba_{2r} \circ \cdots \circ  \ba_{Nr} \triangleq \hat \eX,
\end{equation}
where $R$ is a positive integer and $\ba_{nr} \in \mathbb{R}^{I_n}, n= 1,\ldots, N$, are the factor vectors for the $r$-th component. The minimal number of rank-one components required to represent $\eX$ is called the rank of the tensor, denoted by $\text{rank}(\eX)$ \citep{kruskal_1977, hitchcock_1927}. The compact form of the decomposition can be written using factor matrices:
\begin{equation}
    \eX \approx [\![ \cA_1, \cA_2,\ldots, \cA_N]\!],
\end{equation}
where $\cA_n = [\ba_{n1}, \ba_{n2}, \ldots, \ba_{nR}]\in \mathbb{R}^{I_n \times R}, n = 1, 2, \ldots, N$, are the factor matrices containing the corresponding vectors of each rank-one component. Furthermore, letting $\{\tilde \cA_n| n = 1, 2, \ldots, N\}$ represent the factor matrices with columns normalized to norm one and the weights of rank-1 components absorbed into a vector $\bm \lambda =[\lambda_1, \lambda_2, \ldots, \lambda_R]^\top\in \mathbb{R}^R$, we can write:
\begin{equation}
    \eX \approx [\![ \bm\lambda ;\tilde\cA_1, \tilde\cA_2,\ldots, \tilde\cA_N ]\!] = \sum_{r=1}^R \lambda_r \tilde\ba_{1r} \circ \tilde\ba_{2r} \circ \cdots \circ  \tilde\ba_{Nr}.
\end{equation}

To compute the CP decomposition of a tensor, one of the most widely used approaches is the Alternating Least Squares (ALS) algorithm. The CP-ALS algorithm iteratively updates each factor matrix while keeping the others fixed, thereby minimizing the squared Frobenius norm of the residual tensor. The algorithm begins by initializing the factor matrices $\cA_n\in \mathbb{R}^{I_n \times R}$ for $n = 1, \ldots, N$, where $R$ is the prespecified rank of the decomposition. For each iteration, the $n$-th factor matrix $\cA_n$ is updated by solving a linear least squares problem in closed form. This process involves matricizing the tensor along the $n$-th mode and expressing it in terms of the Khatri-Rao product of the remaining factor matrices. The algorithm continues iteratively until convergence is achieved, as determined by the change in fit or the maximum number of iterations. 

\section{Linked Data Factorization: Related Methods}
\subsection{BIDIFAC}
There is an extensive literature on the factorization of linked matrices that share common dimensions. The BIDIFAC approach is one of the methods focusing on joint decomposition of bidimensionally linked matrices \citep{park_lock_2019}. For clarity, consider the example of $K$ matrices $\{\cX_1, \cX_2, \ldots, \cX_K| \cX_k \in \mathbb{R}^{I_0\times I_1^{(k)}}, k = 1,2,\ldots,K\}$ that are horizontally linked, i.e., sharing the same rows (samples). The BIDIFAC method decomposes matrices as
\begin{equation}{\label{eq:2_1}}
    \cX_k = \cA_0^{(\text{share})} \cA_1^{(\text{share}, k)\top} + \cA_0^{(\text{indiv}, k)} \cA_1^{(\text{indiv}, k)\top} + \cE^{(k)} \text{ for } k=1,\ldots, K.
\end{equation}
The common sample loading matrix $\cA_0^{(\text{share})}\in \mathbb{R}^{I_0 \times R_\text{share}}$ represents the shared structures, explaining variability across multiple rows (samples). The score matrices $\cA_1^{(\text{share}, k)} \in \mathbb{R}^{I_1^{(k)} \times R_\text{share}}$ indicate how these loadings are expressed across columns in each dataset. Individual structure is represented by the loading matrices $\cA_0^{(\text{indiv}, k)}\in \mathbb{R}^{I_0 \times R_\text{indiv}^{(k)}}$ and score matrices $\cA_1^{(\text{indiv}, k)} \in \mathbb{R}^{I_1^{(k)} \times R_\text{indiv}^{(k)} }$ specific to each data source. $\cE^{(k)} \in \mathbb{R}^{I_0\times I_1^{(k)}}$ represents the Gaussian error term. $R_\text{share}$ is the rank of shared structures across datasets and $R_\text{indiv}^{(k)}$ is the rank of individual structure for the $k$-th matrix. Model \ref{eq:2_1} can be obtained by solving the following objective function:
\begin{align}
    f(\cA_0^{(\text{share})}, \ & \cA_0^{(\text{indiv}, k)}, \cA_1^{(\text{share}, k)}, \cA_1^{(\text{indiv}, k)}| k =1, \ldots, K) \nonumber\\
    =&\sum_{k=1}^K \| \cX_k -\cA_0^{(\text{share})} \cA_1^{(\text{share}, k)\top} - \cA_0^{(\text{indiv}, k)} \cA_1^{(\text{indiv}, k)\top}  \|_F^2 \nonumber\\
    &+ \sigma \left( \| \cA_0^{(\text{share})} \|_F^2 + \sum_{k=1}^K \| \cA_1^{(\text{share}, k)} \|_F^2\right) + \sum_{k=1}^K \sigma_k\left( \| \cA_0^{(\text{indiv}, k)} \|_F^2 + \|\cA_1^{(\text{indiv}, k)}\|_F^2 \right).
\end{align}

The Frobenius norm penalty on the factor matrices is motivated by its equivalence to the nuclear norm penalty on the corresponding matrix structures when minimizing the objective function. This relationship further motivates the choice of penalty factors $\{\sigma, \sigma_k|k=1,\ldots, K\}$ by utilizing the soft-thresholding property of nuclear norm penalty \citep{mazumder_2010}.

\subsection{Coupled Matrix and Tensor Factorization}
Building on the concept of integrative factorization for multiple matrices, coupled tensor and matrix factorization (CMTF) extends this approach to simultaneously decompose tensors and matrices that are linked in a shared dimension \citep{acar_2011, acar_2013}. This method allows for the extraction of shared latent structures across different data forms. A CMTF problem for a third-order tensor $\eX \in \mathbb{R}^{I_0\times I_1\times I_2}$ and a matrix $\cY \in \mathbb{R}^{I_0\times I_3}$ is typically formulated as 
\begin{equation}
    f(\cA_0, \cA_1, \cA_2, \cA_3) = \|\eX - [\![\cA_0, \cA_1, \cA_2]\!] \|^2 + \|\cY - \cA_0 \cA_3^\top\|^2
\end{equation}
This formula implicitly assumes that all columns of factor matrix $\cA_0$, i.e. $\{\ba_{0r} |r=1,\ldots, R\}$, are fully shared across datasets. To model shared and unshared structure, Acar et al. proposed to impose $L_1$ penalty to promote sparsity\citep{acar_2013}. The modified objective function is shown below:
\begin{align}
    f(\bm\lambda, \bd, \tilde\cA_0, \tilde\cA_1, \tilde\cA_2, \tilde\cA_3) &= \|\eX - [\![ \bm \lambda; \tilde\cA_0, \tilde\cA_1, \tilde\cA_2 ]\!] \|^2 + \|\cY - \tilde\cA_0 \cD \tilde\cA_3^\top\|^2 + \sigma \|\bm\lambda\|_1 + \sigma \|\bd\|_1\\
    &\text{s.t. }\|\ba_{0r}\| = \|\ba_{1r}\| = \|\ba_{2r}\| = \|\ba_{3r}\| = 1 \text{ for } r=1,\ldots, R, \notag
\end{align}
where $\bm \lambda$ and $\bd$ are the corresponding weights of rank-one components in the factorization of tensor and matrix and $\{\tilde\cA_0, \tilde\cA_1, \tilde\cA_2, \tilde\cA_3\}$ are normalized factor matrices with column norms to be 1. By imposing the $L_1$ penalty, the weights are sparsified so that unshared components will have 
$0$ weight in one of the datasets and shared components will have non-zero weight in both datasets.

\subsection{Structured Data Fusion}
The integrative factorization problem in the previous two sections is often referred to as a data fusion problem when analyzing data from multiple sources. Sorber et al. proposed a highly flexible framework, structured data fusion (SDF), to factorize multiple tensors that can be linked in various ways \citep{sorber_sdf_2015}. This general method can handle different types of factorization models, incorporate various penalties, and enable the representation of factors as functions of latent variables. Although SDF is versatile and applicable to a wide range of integrative factorization problems, it does not provide clear guidelines for selecting appropriate penalty parameters or determining tensor rank. To address the optimization challenge, the authors proposed two techniques: quasi-Newton and nonlinear least squares. For practical implementation, a MATLAB package called Tensorlab was developed, serving as the main comparison of our proposed model. 

\section{Proposed Methods}

\subsection{Single Tensor Penalized Factorization}
For the CP decomposition of a single tensor $\eX \in \mathbb{R}^{I_1\times \cdots \times I_N}$, we consider an objective function with $L_2$ penalties on the factor matrices as follows:
\begin{align} \label{eq:sing_tensor}
f = \big\|\eX-[\![\cA_{1},\ldots,\cA_N ]\!] \big\|_F^2 + \sigma \sum_{i=1}^N \big\| \cA_i \big\|_F^2,
\end{align}
where $\sigma$ is the penalty factor. Similar to the ALS algorithm for the unpenalized CP decomposition, we can solve the above objective by iteratively updating each factor matrix in a closed form while keeping the other fixed. The details are outlined in Algorithm \ref{alg:1}.

\begin{algorithm}
\caption{Alternating Least Squares (ALS) for CP Decomposition with $L_2$ Penalty}
\label{alg:1}
\begin{algorithmic}[1]
    \STATE \textbf{Input:} Tensor $\eX$, rank $R$, penalty factor $\sigma$, maximum iterations $T$, tolerance $\epsilon$
    \STATE \textbf{Output:} Factor matrices $\mathbf{A}_1, \ldots, \mathbf{A}_N$
    \STATE Initialize factor matrices $\mathbf{A}_1, \ldots, \mathbf{A}_N$ randomly
    \REPEAT
        \FOR{$i = 1, \ldots, N$}
        \STATE Update the $i$-th factor matrices as
            $$
            \mathbf{A}_i \leftarrow \arg \min_{\mathbf{A}_i} \left\| \cX_{(i)} - \mathbf{A}_i \left( \bigodot_{\substack{j=N \\ j \neq i}}^1 \mathbf{A}_j \right)^T \right\|_F^2 + \sigma \left\| \mathbf{A}_i \right\|_F^2,
            $$
            where $\cX_{(i)}$ is the mode-$i$ matricization of the tensor $\eX$. This update has a closed-form solution similar to the ridge regression framework \citep{hoerl_kennard_1970_ridge}.
        \ENDFOR
    \UNTIL improvement of objective function is less than $\epsilon$ or maximum iteration $T$ is reached.
\end{algorithmic}
\end{algorithm}
The following Theorem \ref{thm:1} extends Proposition~
\ref{prop:2} to the tensor context, demonstrating that a Frobenius norm penalty on the factor matrices in a tensor is equivalent to a penalty on the component weights (analogous to singular values) for the tensor.  

\begin{theorem}\label{thm:1}
For the CP decomposition problem of a $N$-way tensor, a $L_2$ penalty on factor matrices $\cA_i$, $i=1,\ldots,N$ is equivalent to a $L_{2/N}$ penalty on weights $\bm \lambda$, i.e,
$$
\operatorname*{min}\big\|\eX-[\![\cA_{1},\ldots,\cA_N ]\!] \big\|_F^2 + \sigma \sum_{i=1}^N \big\| \cA_i \big\|_F^2 = \operatorname*{min}\big\|\eX-[\![\bm\lambda; \tilde\cA_{1},\ldots,\tilde\cA_N ]\!] \big\|_F^2 +\sigma N \|\bm \lambda\|_{2/N}^{2/N},
$$
where $\{\tilde\cA_{1},\ldots,\tilde\cA_N\}$ are normalized factor matrices with column norms to be 1.
\end{theorem}
For any tensor order $N>2$, $L_{2/N}$ is a sparsity-inducing penalty and thus can shrink rank-1 components to $0$. This result utilizes the established property of the $L_p$ penalty in penalized regression models, where applying an $L_p$ penalty with $p \leq 1$ is known to shrink coefficients to zero \citep{tibshirani_1996_lasso, fu_1998}.  Theorem \ref{thm:2} extends this result to show how the adjusted weights can be calculated from the unpenalized decomposition in a rank-1 approximation problem. 

\begin{theorem}\label{thm:2}
Suppose that $(\hat{\lambda}, \hat{\ba}_1, \hat{\ba}_2, \ldots, \hat{\ba}_N)$ is the solution to the rank-1 approximation problem for a N-order tensor $\eX$.
\begin{equation*}
    \operatorname*{min} \big \| \eX - [\![ \lambda; \ba_1, \ba_2, \ldots, \ba_N]\!] \big \|_F^2
\end{equation*}
Then the solution to the penalized approximation problem
\begin{equation*}
    \operatorname*{min} \big \| \eX - [\![ \lambda; \ba_1, \ba_2, \ldots, \ba_N]\!] \big \|_F^2 + N\sigma \|\lambda\|_{2/N}^{2/N}
\end{equation*}
will be $[\![\hat{\lambda}_p; \hat{\ba}_1, \hat{\ba}_2, \ldots, \hat{\ba}_N]\!]$, where $\hat{\lambda}_p=\operatorname*{argmin}_{\lambda} (\lambda - \hat{\lambda})^2 + N\sigma \lambda^{2/N}$.
\end{theorem}

By setting $N=2$, the result reduces to $\hat{\lambda}_p = \hat{\lambda} - \sigma$, which is consistent with the soft-thresholding result for a matrix in Proposition~\ref{prop:1}.  An analogous result does not hold for rank greater than $1$, i.e., the factor matrices $\hat{\cA}_i$  will not necessarily be proportional to that for the unpenalized decomposition.     

\subsection{Linked Tensor Factorization}
Building upon the foundations of single tensor decomposition and the integrative factorization of matrices, we extend these techniques to the joint analysis of multiple tensors. The concept of ``linked" matrices can be naturally generalized to the tensor setting. We define multiple tensors as ``linked" if each pair shares the same number of dimensions in at least one mode. Specifically, a set of $K$ one-dimensionally linked tensors is defined as multiple tensors of different orders that share a common first mode (without loss of generality, we can refer to the shared mode as the first mode): 
$\{\eX_k:I_0\times I_1^{(k)}\times\cdots\times I_{N_k}^{(k)}|k=1,\ldots,K\}$. The proposed tensor decomposition formula is defined as
\begin{equation}
\begin{aligned}
    \eX_k\approx[\![ \bm \lambda_k; \tilde\cA_0, \tilde\cA_1^{(k)},\ldots,\tilde\cA_{N_k}^{(k)} ]\!], k = 1, \ldots, K.
\end{aligned}
\end{equation}

Instead of modeling shared and individual structures explicitly like BIDIFAC or adding an $L_1$ penalty on component weights like CMTF, we choose to model the shared and individual structures by $L_2$ penalization on the factor matrices as in \eqref{eq:sing_tensor}:
\begin{equation}\label{eq:9}
        f = \sum_{k=1}^K\big\|\eX_k-[\![\cA_0, \cA_1^{(k)},\ldots,\cA_{N_k}^{(k)} ]\!] \big\|_F^2  +\sigma\left(\big\|\cA_0\big\|_F^2 + \sum_{k=1}^K\sum_{i=1}^{N_k} \big\|\cA_i^{(k)}\big\|_F^2\right).
    \end{equation}
This objective is derived from Theorem \ref{thm:3}, which extends the rank sparsity property from the decomposition of a single tensor to the decomposition of multiple tensors. Theorem \ref{thm:3} demonstrates that applying an $L_2$ penalty to the factor matrices induces sparsity by shrinking some of the component weights to zero. In this framework, each tensor is initially assumed to be approximated by the sum of $R$ rank-1 tensors, where $R$ is a pre-specified initial rank. However, due to rank sparsity, the effective rank of each tensor becomes smaller, with each tensor ultimately approximated by the sum of only a subset of these rank-1 tensors. Shared components retain non-zero weights across all datasets, while individual components have non-zero weights exclusively in their respective datasets.

\begin{theorem}\label{thm:3}
The minimization problem for objective function \eqref{eq:9} is equivalent to the following problem with mixed $L_p$ and $L_2$ penalty:
\begin{equation*}\label{12}
\begin{aligned}
    &\operatorname{min} \sum_{k=1}^K\big\|\eX_k-[\![\cA_0, \cA_1^{(k)},\ldots,\cA_{N_k}^{(k)} ]\!] \big\|_F^2  +\sigma\left(\big\|\cA_0\big\|_F^2 + \sum_{k=1}^K\sum_{i=1}^{N_k} \big\|\cA_i^{(k)}\big\|_F^2\right)\\
    =\ &\operatorname{min}\sum_{k=1}^K\big\|\bm \eX_k-[\![\bm \lambda_0* \bm \lambda_{(0)k}; \tilde\cA_0, \tilde\cA_1^{(k)},\ldots,\tilde\cA_{N_k}^{(k)} ]\!]\big\|_F^2 + \sigma\left( \|\bm\lambda_0\|_2^2 + \sum_{k=1}^K N_k\|\bm\lambda_{(0)k} \|_{2/N_k}^{2/N_k}\right),
\end{aligned}
\end{equation*}
where the $r$-th element of $\bm \lambda_0$ is the Frobenius norm for the $r$-th column of shared factor matrix $\cA_0$. Similarly, the $r$-th element of $\bm \lambda_{(0)i}$ is the product of Frobenius norms for the $r$-th column of remaining factor matrices $\cA_{i}^{(k)}$, for $i=1,\ldots, N_k$.
\end{theorem}

\begin{figure}[!h]
    \centering
    \subfigure[MULTIFAC Model]{\includegraphics[width=0.7\linewidth]{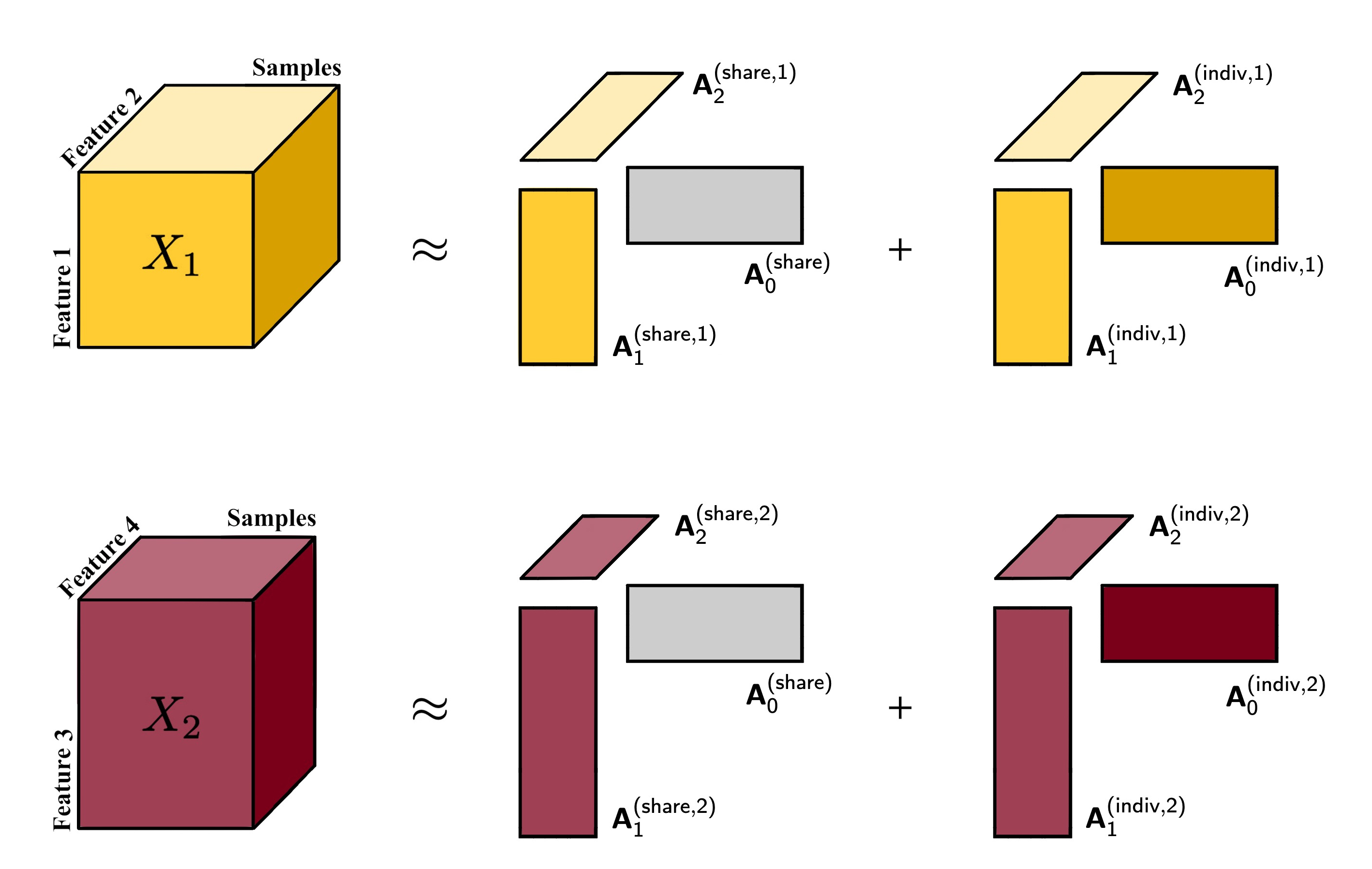}}

    \subfigure[Rank Sparsity Property]{\includegraphics[width=0.95\linewidth]{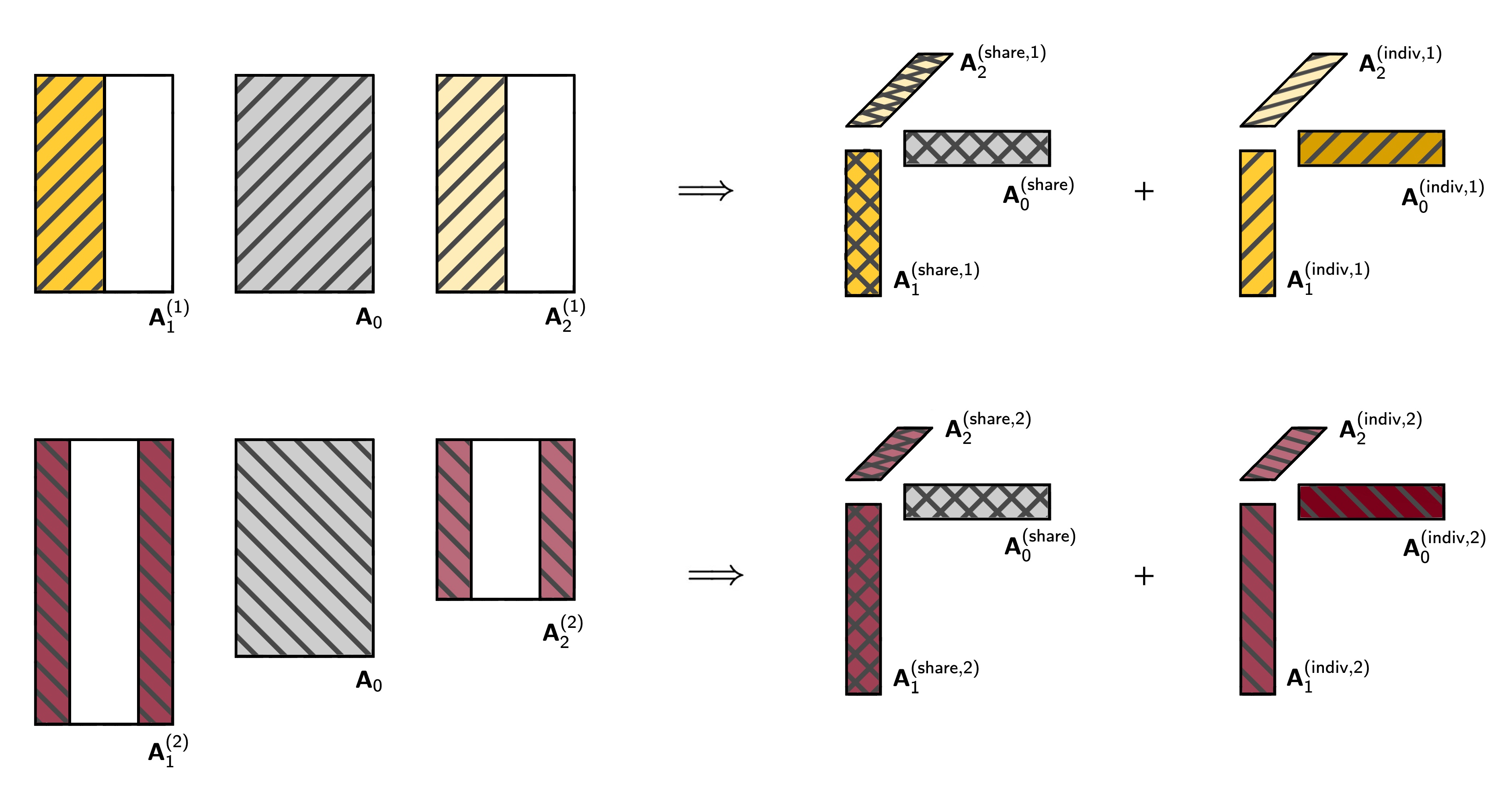}}
    
    \caption{An illustration of MULTIFAC for two linked 3-way tensors. Subfigure (a) demonstrates how MULTIFAC decomposes tensors into shared and individual structures, where the individual components of the two tensors have distinct factor matrices $\cA_0^{(\text{indiv, 1})}$ and $\cA_0^{(\text{indiv, 2})}$. In practice, however, these individual structures are not explicitly modeled; instead, both tensors are assumed to share the same $\cA_0$, as depicted in the left-hand side of subfigure (b). Through penalization, certain columns are shrunk to zero, indicated by white regions, effectively reducing the rank and leading to the emergence of individual structures.}
    \label{fig:multifac}
\end{figure}

An advantage of using an $L_2$ penalty on the factor matrices, compared to an $L_1$ norm on the weights $\bm \lambda$, is that our proposed objective function \eqref{eq:9} remains smooth while still retaining the sparsity-inducing properties of the $L_1$ penalty. This smoothness facilitates straightforward optimization, allowing for simple modifications to Algorithm \ref{alg:1}. With penalty factor $\sigma$ fixed, the ALS algorithm can be easily extended to handle multiple linked tensor decomposition as detailed in Algorithm \ref{alg:2}. Another advantage is that this objective leads to automatic rank selection with a single penalty. We begin with an upper bound on the overall rank $R$ ($\cA_{i}^{(k)}: I_i^{(k) \times R)}$), and the sparsity-inducing property of the penalty leads to lower rank shared and unshared structures as detailed in Figure~\ref{fig:multifac}. 

\begin{algorithm}[!h]
\caption{Alternating Least Squares (ALS) for Multi-Tensor CP Decomposition with $L_2$ Penalty}
\label{alg:2}
\begin{algorithmic}[1]
    \STATE \textbf{Input:} Tensor $\{\eX_k|k=1,\ldots,K\}$, rank $R$, penalty factor $\sigma$, maximum iterations $T$, tolerance $\epsilon$
    \STATE \textbf{Output:} Factor matrices $\cA_0$, $\{\cA_1^{(k)},\ldots,\cA_{N_k}^{(k)}| k=1, \ldots,K\}$.
    \STATE Initialize factor matrices randomly
    \REPEAT
        \STATE Update shared factor matrices as
        \begin{equation*}
            \cA_0 \leftarrow \arg \min_{\cA_0} \sum_{k=1}^K \left\| \cX_{k(1)} - \cA_0 \left( \bigodot_{j=N_k}^1 \cA_j^{(k)} \right)^T \right\|_F^2 + \sigma \left\| \cA_0 \right\|_F^2,
        \end{equation*}  
        where the update for $\cA_0$ has a closed-form solution similar to ridge regression \citep{hoerl_kennard_1970_ridge}.
        \FOR{$k = 1, \ldots, K$}
            \FOR{$i = 1, \ldots, N_k$}
                \STATE Update individual factor matrices as
                \begin{equation*}
                    \cA_i^{(k)} \leftarrow \arg \min_{\cA_i^{(k)}} \left\| \cX_{k(i+1)} - \cA_i^{(k)} \left( \bigodot_{\substack{j=N_k \\ j \neq i}}^1 \cA_j^{(k)} \odot \cA_0 \right)^T \right\|_F^2 + \sigma \left\| \cA_i^{(k)} \right\|_F^2,
                \end{equation*} 
                where the update for $\cA_i^{(k)}$ has a closed-form solution similar to ridge regression \citep{hoerl_kennard_1970_ridge}.
            \ENDFOR
        \ENDFOR
    \UNTIL improvement of objective function is less than $\epsilon$ or maximum iteration $T$ is reached.
\end{algorithmic}
\end{algorithm}

\subsection{Missing Data  Imputation}\label{sec: GALS}

When dealing with missing data, this algorithm can be extended using an expectation-maximization (EM) approach. During each iteration, the missing entries are updated based on their estimated low-rank approximations. Similar algorithms have been used for imputation of a single tensor with a CP factorization \cite{acar2011scalable}. 
 We consider two types of missing patterns. The first type is tensor-wise missing, where data is missing in an entire sample for one of the $K$ tensors, meaning all entries in a particular mode are absent. The second type is entry-wise missing, where individual entries are randomly missing throughout the tensor.

\begin{algorithm}

\caption{EM-ALS for Multi-Tensor CP Decomposition with $L_2$ Penalty}
\label{alg:3}
    \begin{algorithmic}[1]
        \STATE \textbf{Input:} Tensor $\{\eX_k|k=1,\ldots,K\}$, rank $R$, penalty factor $\sigma$, maximum iterations $T$, tolerance $\epsilon$
        \STATE \textbf{Output:} Factor matrices $\cA_0$, $\{\cA_1^{(k)},\ldots,\cA_{N_k}^{(k)}| k=1, \ldots,K\}$.
        \STATE Initialize factor matrices randomly.
        \STATE Impute entry-wise missing values with current low-rank structure and impute tensor-wise missing values with shared low-rank structure.
        \STATE Update factor matrices as in Step 5 to 10 in Algorithm \ref{alg:2}.
        \STATE Repeat Step 4 and 5 until convergence.
        \STATE Impute entry-wise missing values with current low-rank structure and impute tensor-wise missing values with shared low-rank structure.
    \end{algorithmic}
\end{algorithm}

During the iterative imputation process, the entry-wise missing values are updated using their corresponding low-rank approximations derived from the current estimates. For tensor-wise missing data, only the shared low-rank structures across samples are used for updating. This approach is adopted because other samples do not provide complete information about an entirely missing sample, particularly regarding the scale (i.e., sample loading) of a completely missing sample. However, the shared structures can still offer valuable scale information by leveraging the available data sample from other sources to help impute the shared sample loading. The imputation algorithm is described in Algorithm \ref{alg:3}.

\subsection{Parameter Tuning and Rank Selection}
\label{sec: two_step_estimation}
In this section, we discuss the estimation procedure for both parameter tuning and the selection of the tensor ranks. The proposed framework consists of two cross-validation steps: step 1 determines the tensor ranks using $L_2$ penalization, while step 2 further tunes the penalty parameter $\sigma$ with the ranks fixed. A key summary statistic employed in this process is the relative squared error (RSE), defined as:
$$
\text{RSE}(\hat{\eX},\eS) = \frac{\|\hat\eX - \eS\|_F^2}{\|\eS\|_F^2},
$$
where $\hat{\eX}$ represents the estimated structure and $\eS$ is the true signal $\eX$. In practice the true low-rank signal is unknown, and in what follows we use the RSE for data artificially held-out as missing, that is, $\text{RSE}_{\text{missing}}=\text{RSE}(\hat{X}_\text{missing},X_\text{missing})$.   

Step 1: Determine tensor rank using cross-validation. The first cross-validation step addresses the challenge of determining the ranks of the underlying shared and unshared signals. Starting with a large initial rank, the rank sparsity property is leveraged to guide the selection process. Cross-validation is  performed, in which data are artificially held out as missing and imputed, over different penalty factors $\sigma$.  The performance metric is the average $\text{RSE}_{\text{missing}}$ calculated over the validation sets. We employ the one standard error rule \citep{1se_rule}. This rule selects the most parsimonious model whose performance is within one standard error of the minimum $\text{RSE}_{\text{missing}}$ observed across the grid. The rationale behind this approach is to favor simpler models that are still close in performance to the best model, thereby avoiding overfitting and enhancing generalizability. The ranks (i.e. number of non-zero components) for the shared and unshared structures in the selected model will be fixed for Step 2. 

Step 2: Further tune the penalty parameter $\sigma$. In the second cross-validation step, we fix the tensor ranks as determined in the first step and perform a grid search to fine-tune the penalty parameter $\sigma$. In this step, we aim to identify the optimal $\sigma$ that yields the best imputation performance, thereby ensuring the most accurate factorization. The grid search evaluates a range of penalty values, and the optimal $\sigma$ is selected as the one that minimizes the average $\text{RSE}_{\text{missing}}$  across the validation sets.

There are several important practical considerations when calculating the decomposition for multiple tensors. Previous studies have shown that the ALS algorithm is not guaranteed to converge to a global minimum and can be significantly influenced by the choice of initialization \citep{kolda_tensor_review_2009}. To mitigate this issue, we begin with multiple randomly initialized factor matrices and select the solution that achieves the lowest value of the final unpenalized objective function. Additionally, to stabilize the results, we employ a technique known as tempered regularization, where we start with a low penalty value and gradually increase it to the desired level \citep{lock_park_2020}. This approach facilitates a smoother optimization process toward the optimal solution.

\section{Simulation}
To evaluate the performance of our proposed tensor decomposition algorithm, we conducted simulation studies for both single and multiple tensor decomposition. These simulations assess the accuracy and robustness of the algorithm under various noise conditions and dimensions. The tensor data consist of a true low-rank signal and additive noise. For all simulations, the true signal was generated using factor matrices, where each column was sampled from a standard normal distribution. Noise was simulated from a normal distribution, with variance adjusted to achieve predefined signal-to-noise ratio (SNR) levels. To evaluate robustness under varying noise conditions, the signal-to-noise ratio (SNR) was set at three levels: $1/3$, 1, and 3 across all simulation settings.

\subsection{Single Tensor Decomposition}
\subsubsection{Simulation Setup}
We first evaluate the method's performance on single tensor decomposition before extending the analysis to multiple tensor decomposition. The generated tensor has size $50\times 50\times 50$. The low-rank for the tensor was set to be 5 and the pre-specified rank in estimation under Algorithm!~\ref{alg:2} was set to be 20. Performance metrics were calculated at each step of the selection procedure in section~\ref{sec: two_step_estimation}, including the parsimonious model with selected $\sigma$ value at step 1, unpenalized model with selected rank after step 1, and the ``optimal" model after step 2. We also compared them with the standard unpenalized  CP decomposition model with true rank. 

We considered the decomposition of both complete tensors and tensors with missing data. For the complete tensor scenario, we evaluated the decomposition accuracy of our methods and compared it with the Nonlinear Least Squares (NLS) solver given true rank from the Tensorlab package \citep{sorber_sdf_2015}. In the missing data imputation simulation, we randomly removed 10\% of each tensor and assess the imputation performance. Each simulation setting was repeated 100 times to assess the stability of the results.
\subsubsection{Simulation Results}
Table \ref{table:2.5} presents the decomposition accuracy for single complete tensor decomposition, measured using RSE across different SNR levels. Notably, once the rank was selected in our model, there was no significant difference in decomposition performance across methods. It is worth noticed that our method consistently picked the true rank in step 1 of coss validation across all settings.
Table \ref{table:2.6} summarizes the results for single tensor imputation, measured by RSE comparing imputed value with underlying missing signal. The results indicate that our ``optimal" model consistently outperforms alternative methods, even when the true rank is specified, with the performance gap becoming more pronounced as the SNR increases. 

\begin{table}[!h]
\centering
\caption{Relative square error for single complete tensor decomposition}
\begin{tabular}{|l|ll|ll|ll|}
\hline
 & \multicolumn{2}{l|}{SNR = 1/3} & \multicolumn{2}{l|}{SNR = 1} & \multicolumn{2}{l|}{SNR = 3} \\ \hline
\textbf{Method} & \multicolumn{1}{l|}{Mean} & SD & \multicolumn{1}{l|}{Mean} & SD & \multicolumn{1}{l|}{Mean} & SD \\ \hline
Step 1 & \multicolumn{1}{l|}{0.1588} & 0.008 & \multicolumn{1}{l|}{0.0958} & 0.0056 & \multicolumn{1}{l|}{0.0578} & 0.0037 \\ \hline
Constraint & \multicolumn{1}{l|}{0.134} & 0.0038 & \multicolumn{1}{l|}{0.0769} & 0.0021 & \multicolumn{1}{l|}{0.0443} & 0.0012 \\ \hline
Step 2 & \multicolumn{1}{l|}{0.1336} & 0.0038 & \multicolumn{1}{l|}{0.0768} & 0.0021 & \multicolumn{1}{l|}{0.0443} & 0.0012 \\ \hline
True Rank & \multicolumn{1}{l|}{0.134} & 0.0038 & \multicolumn{1}{l|}{0.0769} & 0.0021 & \multicolumn{1}{l|}{0.0443} & 0.0012 \\ \hline
NLS & \multicolumn{1}{l|}{0.134} & 0.0038 & \multicolumn{1}{l|}{0.0769} & 0.0021 & \multicolumn{1}{l|}{0.0443} & 0.0012 \\ \hline
\end{tabular}
\label{table:2.5}
\end{table}

\begin{table}[!h]
\centering
\caption{Relative square error for single tensor imputation}
\begin{tabular}{|l|ll|ll|ll|}
\hline
 & \multicolumn{2}{l|}{SNR = 1/3} & \multicolumn{2}{l|}{SNR = 1} & \multicolumn{2}{l|}{SNR = 3} \\ \hline
Mehod & \multicolumn{1}{l|}{Mean} & SD & \multicolumn{1}{l|}{Mean} & SD & \multicolumn{1}{l|}{Mean} & SD \\ \hline
Step 1 & \multicolumn{1}{l|}{0.1653} & 0.0067 & \multicolumn{1}{l|}{0.101} & 0.0045 & \multicolumn{1}{l|}{0.0611} & 0.0034 \\ \hline
Constraint & \multicolumn{1}{l|}{0.1419} & 0.0044 & \multicolumn{1}{l|}{0.0901} & 0.0515 & \multicolumn{1}{l|}{0.0729} & 0.0893 \\ \hline
Step 2 & \multicolumn{1}{l|}{0.1405} & 0.0043 & \multicolumn{1}{l|}{0.0811} & 0.0024 & \multicolumn{1}{l|}{0.0468} & 0.0014 \\ \hline
True Rank & \multicolumn{1}{l|}{0.1419} & 0.0044 & \multicolumn{1}{l|}{0.0892} & 0.0462 & \multicolumn{1}{l|}{0.0715} & 0.0851 \\ \hline
\end{tabular}
\label{table:2.6}
\end{table}

\subsection{Multiple Tensor Decomposition}
\subsubsection{Simulation Setup}
In this section, we conducted two sets of simulations to evaluate the performance of the MULTIFAC method in the context of linked tensor factorization. Specifically, we assess (1) the factorization of multiple complete tensors and (2)the imputation of missing values. Each simulation was performed using two different tensor dimension settings:
\begin{itemize}
    \item Tensors with identical dimensions $50 \times 50 \times 50$;
    \item Tensors with different dimensions: the first tensor is $100 \times 100 \times 4$ and the second tensor is $100 \times 40 \times 10 \times 3$,
\end{itemize}
where the first mode of the tensors was assumed to be linked in both settings. The true rank of the tensors was set as 2 for shared structures and 3 for individual structures in each tensor. Each scenario was repeated 50 times.

For the first simulation with complete tensors, we compared the  decomposition accuracy of MULTIFAC and the NLS solver from the Tensorlab package \citep{sorber_sdf_2015}. For NLS, we used two rank settings: the true rank and twice the true rank. The RSEs were computed to evaluate the accuracy in recovering the full signal, shared components, and individual components. In the second simulation, we evaluated the imputation performance of MULTIFAC when tensors contain missing values. For each tensor, 10\% of the entries were randomly set to be missing, comprising 5\% tensor-wise missing entries and 5\% entry-wise missing entries. We reported the RSE for all missing entries, as well as separately for randomly missing entries and tensor-wise missing entries.

\subsubsection{Simulation Results}
Tables \ref{table:2.1} and \ref{table:2.2} present the decomposition accuracy for complete tensors across various SNR levels. The pre-specified rank for MULTIFAC was set to 20, exceeding the true tensor rank. Overall, MULTIFAC consistently outperformed NLS in terms of RSE under all tested conditions, demonstrating greater accuracy in uncovering the underlying signals. When comparing the ability to decompose different structures, MULTIFAC significantly outperformed NLS. Notably, NLS often failed to correctly estimate these structures (SNR $>$ 1), even when the true rank was provided.

Imputation performance is summarized in Tables \ref{table:2.3} and \ref{table:2.4}. The RSE for entry-wise imputation closely matched the RSE for the observed tensor entries. As expected, tensor-wise imputation was less accurate than entry-wise imputation, since only shared structures were used for imputing missing tensors. Nonetheless, with RSE values significantly below 1, it can be concluded that MULTIFAC achieved a reasonable degree of accuracy in recovering tensor-wise missing entries.

\begin{table}[!h]
\caption{Performance metrics for decomposing complete tensors of varying sizes. The superscripts denote the tensor being analyzed (e.g., $\text{RSE}^1_\text{full}$ for Tensor 1) while subscripts identify the structural components evaluated: ``full" for the complete signal, ``share" for shared signals across tensors, and ``indiv" for individual tensor signals. Each cell reports the average RSEs, with the standard deviation of RSEs shown in parentheses.}
\renewcommand{\arraystretch}{1.2}
\centering
\begin{tabular}{|c|c|c|c|c|c|c|c|}
\hline
Method & SNR &$\text{RSE}^1_\text{full}$& $\text{RSE}^2_\text{full}$& $\text{RSE}^1_\text{share}$ & $\text{RSE}^2_\text{share}$ & $\text{RSE}^1_\text{Indiv}$ & $\text{RSE}^2_\text{Indiv}$ \\ \hline
MULTIFAC & \multirow{5}{*}{3} & \begin{tabular}[c]{@{}c@{}}0.113\\ (0.025)\end{tabular} & \begin{tabular}[c]{@{}c@{}}0.046\\ (0.003)\end{tabular} & \begin{tabular}[c]{@{}c@{}}0.162\\ (0.187)\end{tabular} & \begin{tabular}[c]{@{}c@{}}0.112\\ (0.227)\end{tabular} & \begin{tabular}[c]{@{}c@{}}0.156\\ (0.095)\end{tabular} & \begin{tabular}[c]{@{}c@{}}0.116\\ (0.233)\end{tabular} \\ 
NLS with true rank &  & \begin{tabular}[c]{@{}c@{}}0.25\\ (0.145)\end{tabular} & \begin{tabular}[c]{@{}c@{}}0.212\\ (0.127)\end{tabular} & \begin{tabular}[c]{@{}c@{}}1.327\\ (2.371)\end{tabular} & \begin{tabular}[c]{@{}c@{}}0.975\\ (0.524)\end{tabular} & \begin{tabular}[c]{@{}c@{}}1.028\\ (1.573)\end{tabular} & \begin{tabular}[c]{@{}c@{}}0.814\\ (0.409)\end{tabular} \\ 
NLS with double rank &  & \begin{tabular}[c]{@{}c@{}}0.142\\ (0.008)\end{tabular} & \begin{tabular}[c]{@{}c@{}}0.065\\ (0.011)\end{tabular} & \begin{tabular}[c]{@{}c@{}}1.334\\ (1.765)\end{tabular} & \begin{tabular}[c]{@{}c@{}}1.096\\ (0.672)\end{tabular} & \begin{tabular}[c]{@{}c@{}}1.012\\ (1.003)\end{tabular} & \begin{tabular}[c]{@{}c@{}}0.876\\ (0.455)\end{tabular} \\ \hline
MULTIFAC & \multirow{5}{*}{1} & \begin{tabular}[c]{@{}c@{}}0.169\\ (0.02)\end{tabular} & \begin{tabular}[c]{@{}c@{}}0.083\\ (0.02)\end{tabular} & \begin{tabular}[c]{@{}c@{}}0.235\\ (0.224)\end{tabular} & \begin{tabular}[c]{@{}c@{}}0.144\\ (0.213)\end{tabular} & \begin{tabular}[c]{@{}c@{}}0.232\\ (0.143)\end{tabular} & \begin{tabular}[c]{@{}c@{}}0.161\\ (0.229)\end{tabular} \\ 
NLS with true rank &  & \begin{tabular}[c]{@{}c@{}}0.295\\ (0.117)\end{tabular} & \begin{tabular}[c]{@{}c@{}}0.24\\ (0.13)\end{tabular} & \begin{tabular}[c]{@{}c@{}}1.254\\ (1.211)\end{tabular} & \begin{tabular}[c]{@{}c@{}}1.509\\ (3.305)\end{tabular} & \begin{tabular}[c]{@{}c@{}}1.051\\ (1.093)\end{tabular} & \begin{tabular}[c]{@{}c@{}}1.363\\ (2.941)\end{tabular} \\ 
NLS with double rank &  & \begin{tabular}[c]{@{}c@{}}0.257\\ (0.011)\end{tabular} & \begin{tabular}[c]{@{}c@{}}0.12\\ (0.007)\end{tabular} & \begin{tabular}[c]{@{}c@{}}1.209\\ (0.489)\end{tabular} & \begin{tabular}[c]{@{}c@{}}1.068\\ (0.43)\end{tabular} & \begin{tabular}[c]{@{}c@{}}1.005\\ (0.45)\end{tabular} & \begin{tabular}[c]{@{}c@{}}0.927\\ (0.436)\end{tabular} \\ \hline
MULTIFAC & \multirow{5}{*}{1/3} & \begin{tabular}[c]{@{}c@{}}0.274\\ (0.028)\end{tabular} & \begin{tabular}[c]{@{}c@{}}0.151\\ (0.041)\end{tabular} & \begin{tabular}[c]{@{}c@{}}0.379\\ (0.239)\end{tabular} & \begin{tabular}[c]{@{}c@{}}0.293\\ (0.305)\end{tabular} & \begin{tabular}[c]{@{}c@{}}0.392\\ (0.207)\end{tabular} & \begin{tabular}[c]{@{}c@{}}0.31\\ (0.276)\end{tabular} \\ 
NLS with true rank &  & \begin{tabular}[c]{@{}c@{}}0.361\\ (0.094)\end{tabular} & \begin{tabular}[c]{@{}c@{}}0.262\\ (0.127)\end{tabular} & \begin{tabular}[c]{@{}c@{}}1.378\\ (1.584)\end{tabular} & \begin{tabular}[c]{@{}c@{}}0.903\\ (0.632)\end{tabular} & \begin{tabular}[c]{@{}c@{}}1.141\\ (1.168)\end{tabular} & \begin{tabular}[c]{@{}c@{}}0.761\\ (0.5)\end{tabular} \\ 
NLS with double rank &  & \begin{tabular}[c]{@{}c@{}}0.46\\ (0.019)\end{tabular} & \begin{tabular}[c]{@{}c@{}}0.218\\ (0.016)\end{tabular} & \begin{tabular}[c]{@{}c@{}}1.448\\ (1.177)\end{tabular} & \begin{tabular}[c]{@{}c@{}}0.882\\ (0.464)\end{tabular} & \begin{tabular}[c]{@{}c@{}}1.227\\ (0.908)\end{tabular} & \begin{tabular}[c]{@{}c@{}}0.811\\ (0.492)\end{tabular} \\ \hline
\end{tabular}
\label{table:2.1}
\end{table}

\begin{table}[!h]
\caption{Performance metrics for decomposing complete tensors of same sizes. The superscripts denote the tensor being analyzed (e.g., $\text{RSE}^1_\text{full}$ for Tensor 1) while subscripts identify the structural components evaluated: ``full" for the complete signal, ``share" for shared signals across tensors, and ``indiv" for individual tensor signals. Each cell reports the average RSEs, with the standard deviation of RSEs shown in parentheses.}
\renewcommand{\arraystretch}{1.2}
\centering
\begin{tabular}{|c|c|c|c|c|c|c|c|}
\hline
Method & SNR &$\text{RSE}^1_\text{full}$ & $\text{RSE}^2_\text{full}$ & $\text{RSE}^1_\text{share}$ & $\text{RSE}^2_\text{share}$ & $\text{RSE}^1_\text{Indiv}$ & $\text{RSE}^2_\text{Indiv}$ \\ \hline
MULTIFAC & \multirow{5}{*}{3} & \begin{tabular}[c]{@{}c@{}}0.132\\ (0.043)\end{tabular} & \begin{tabular}[c]{@{}c@{}}0.13\\ (0.043)\end{tabular} & \begin{tabular}[c]{@{}c@{}}0.113\\ (0.037)\end{tabular} & \begin{tabular}[c]{@{}c@{}}0.112\\ (0.04)\end{tabular} & \begin{tabular}[c]{@{}c@{}}0.145\\ (0.05)\end{tabular} & \begin{tabular}[c]{@{}c@{}}0.145\\ (0.049)\end{tabular} \\ 

NLS with true rank &  & \begin{tabular}[c]{@{}c@{}}0.079\\ (0.003)\end{tabular} & \begin{tabular}[c]{@{}c@{}}0.074\\ (0.003)\end{tabular} & \begin{tabular}[c]{@{}c@{}}0.95\\ (0.428)\end{tabular} & \begin{tabular}[c]{@{}c@{}}0.915\\ (0.33)\end{tabular} & \begin{tabular}[c]{@{}c@{}}0.8\\ (0.372)\end{tabular} & \begin{tabular}[c]{@{}c@{}}0.758\\ (0.263)\end{tabular} \\ 
NLS with double rank &  & \begin{tabular}[c]{@{}c@{}}0.079\\ (0.003)\end{tabular} & \begin{tabular}[c]{@{}c@{}}0.074\\ (0.003)\end{tabular} & \begin{tabular}[c]{@{}c@{}}0.95\\ (0.428)\end{tabular} & \begin{tabular}[c]{@{}c@{}}0.915\\ (0.33)\end{tabular} & \begin{tabular}[c]{@{}c@{}}0.8\\ (0.372)\end{tabular} & \begin{tabular}[c]{@{}c@{}}0.758\\ (0.263)\end{tabular} \\ \hline
MULTIFAC & \multirow{5}{*}{1} & \begin{tabular}[c]{@{}c@{}}0.147\\ (0.033)\end{tabular} & \begin{tabular}[c]{@{}c@{}}0.147\\ (0.036)\end{tabular} & \begin{tabular}[c]{@{}c@{}}0.169\\ (0.202)\end{tabular} & \begin{tabular}[c]{@{}c@{}}0.163\\ (0.246)\end{tabular} & \begin{tabular}[c]{@{}c@{}}0.19\\ (0.141)\end{tabular} & \begin{tabular}[c]{@{}c@{}}0.18\\ (0.125)\end{tabular} \\ 
NLS with true rank &  & \begin{tabular}[c]{@{}c@{}}0.202\\ (0.152)\end{tabular} & \begin{tabular}[c]{@{}c@{}}0.328\\ (0.139)\end{tabular} & \begin{tabular}[c]{@{}c@{}}1.013\\ (0.65)\end{tabular} & \begin{tabular}[c]{@{}c@{}}0.852\\ (0.55)\end{tabular} & \begin{tabular}[c]{@{}c@{}}0.882\\ (0.58)\end{tabular} & \begin{tabular}[c]{@{}c@{}}0.698\\ (0.676)\end{tabular} \\ 
NLS with double rank &  & \begin{tabular}[c]{@{}c@{}}0.138\\ (0.004)\end{tabular} & \begin{tabular}[c]{@{}c@{}}0.13\\ (0.004)\end{tabular} & \begin{tabular}[c]{@{}c@{}}1.035\\ (0.401)\end{tabular} & \begin{tabular}[c]{@{}c@{}}0.923\\ (0.329)\end{tabular} & \begin{tabular}[c]{@{}c@{}}0.867\\ (0.305)\end{tabular} & \begin{tabular}[c]{@{}c@{}}0.768\\ (0.27)\end{tabular} \\ \hline
MULTIFAC & \multirow{5}{*}{1/3} & \begin{tabular}[c]{@{}c@{}}0.154\\ (0.015)\end{tabular} & \begin{tabular}[c]{@{}c@{}}0.154\\ (0.016)\end{tabular} & \begin{tabular}[c]{@{}c@{}}0.946\\ (0.363)\end{tabular} & \begin{tabular}[c]{@{}c@{}}0.919\\ (0.423)\end{tabular} & \begin{tabular}[c]{@{}c@{}}0.825\\ (0.267)\end{tabular} & \begin{tabular}[c]{@{}c@{}}0.793\\ (0.29)\end{tabular} \\ 
NLS with true rank &  & \begin{tabular}[c]{@{}c@{}}0.277\\ (0.151)\end{tabular} & \begin{tabular}[c]{@{}c@{}}0.381\\ (0.124)\end{tabular} & \begin{tabular}[c]{@{}c@{}}1.026\\ (0.398)\end{tabular} & \begin{tabular}[c]{@{}c@{}}1.558\\ (2.343)\end{tabular} & \begin{tabular}[c]{@{}c@{}}0.881\\ (0.35)\end{tabular} & \begin{tabular}[c]{@{}c@{}}1.326\\ (1.973)\end{tabular} \\ 
NLS with double rank &  & \begin{tabular}[c]{@{}c@{}}0.241\\ (0.006)\end{tabular} & \begin{tabular}[c]{@{}c@{}}0.226\\ (0.008)\end{tabular} & \begin{tabular}[c]{@{}c@{}}1.208\\ (0.539)\end{tabular} & \begin{tabular}[c]{@{}c@{}}1.075\\ (0.385)\end{tabular} & \begin{tabular}[c]{@{}c@{}}1.003\\ (0.389)\end{tabular} & \begin{tabular}[c]{@{}c@{}}0.92\\ (0.359)\end{tabular} \\ \hline
\end{tabular}
\label{table:2.2}
\end{table}

\begin{table}[!h]
\caption{Imputation performance for tensors of varying sizes. The subscript denote the signal subsets evaluated: ``observe" for observed signal, ``missing" for full missing signal, ``entry-wise" for entry-wise missing signal and ``tensor-wise" for tensor-wise missing signal. Each cell reports the average RSE, with the standard deviation of RSE shown in parentheses.}
\centering
\begin{tabular}{|c|c|c|c|c|c|}
\hline
Tensor Size & SNR & $\text{RSE}_\text{observe}$ & $\text{RSE}_\text{missing}$ & $\text{RSE}_\text{entry-wise}$ & $\text{RSE}_\text{tensor-wise}$ \\ \hline
$100 \times 100 \times 4$ & \multirow{2}{*}{3} & 0.126 (0.032) & 0.543 (0.095) & 0.129 (0.033) & 0.262 (0.206) \\  
$100\times 40 \times 10 \times 3$ &  & 0.049 (0.004) & 0.525 (0.117) & 0.049 (0.004) & 0.232 (0.252) \\ \hline
$100 \times 100 \times 4$ & \multirow{2}{*}{1} & 0.174 (0.016) & 0.564 (0.093) & 0.179 (0.017) & 0.327 (0.221) \\  
$100\times 40 \times 10 \times 3$ &  & 0.084 (0.018) & 0.531 (0.112) & 0.086 (0.019) & 0.281 (0.24) \\ \hline
$100 \times 100 \times 4$ & \multirow{2}{*}{1/3} & 0.286 (0.033) & 0.602 (0.086) & 0.292 (0.036) & 0.447 (0.227) \\  
$100\times 40 \times 10 \times 3$ &  & 0.164 (0.051) & 0.561 (0.112) & 0.166 (0.053) & 0.39 (0.232) \\ \hline
\end{tabular}
\label{table:2.3}
\end{table}

\begin{table}[!h]
\caption{Imputation performance for tensors of same sizes. The subscript denote the signal subsets evaluated: ``observe" for observed signal, ``missing" for full missing signal, ``entry-wise" for entry-wise missing signal and ``tensor-wise" for tensor-wise missing signal. Each cell reports the average RSE, with the standard deviation of RSE shown in parentheses.}
\centering
\begin{tabular}{|c|c|c|c|c|c|}
\hline
Tensor Size & SNR & $\text{RSE}_\text{observe}$ & $\text{RSE}_\text{missing}$ & $\text{RSE}_\text{entry-wise}$ & $\text{RSE}_\text{tensor-wise}$  \\ \hline
$50 \times 50 \times 50$ & \multirow{2}{*}{3} & 0.122 (0.053) & 0.593 (0.103) & 0.124 (0.053) & 0.455 (0.315) \\  
$50 \times 50 \times 50$ &  & 0.12 (0.052) & 0.587 (0.1) & 0.122 (0.053) & 0.445 (0.294) \\ \hline
$50 \times 50 \times 50$ & \multirow{2}{*}{1} & 0.15 (0.033) & 0.555 (0.097) & 0.152 (0.033) & 0.284 (0.16) \\  
$50 \times 50 \times 50$ &  & 0.148 (0.03) & 0.556 (0.104) & 0.15 (0.031) & 0.293 (0.187) \\ \hline
$50 \times 50 \times 50$ & \multirow{2}{*}{1/3} & 0.172 (0.03) & 0.552 (0.095) & 0.174 (0.03) & 0.279 (0.121) \\  
$50 \times 50 \times 50$ &  & 0.171 (0.024) & 0.554 (0.101) & 0.173 (0.025) & 0.288 (0.147) \\ \hline
\end{tabular}
\label{table:2.4}
\end{table}

\section{Real Data Application}
\subsection{Sysmex Hematology and MRI data}
We applied our method to analyze the progression of iron deficiency in a cohort of infant monkeys observed from birth to eight months of age. Hematology indices were collected at 2 weeks, and 2, 4, 6, 8 months after birth using a Sysmex hematology analyzer; for more detail see \citet{rao_lock_etal_2023}. Magnetic Resonance Imaging (MRI) data were also collected after eight months, focusing on four diffusion tensor imaging (DTI) parameters—axial diffusivity (AD), fractional anisotropy (FA), mean diffusivity (MD), and radial diffusivity (RD)—across 14 brain regions. Integrative analysis was conducted using the subset of infants with matched identifiers in both the hematology and MRI datasets. 

The final datasets contains 19 infants, among which 8 developed anemia, defined as hemoglobin levels below 10 g/dL during the observation period. The first dataset is a 3-way tensor contains the hematology information $\eX_1 \in \mathbb{R}^{19\times 5 \times 18}$, organized as $\textit{Monkey Infants} \times \textit{Age} \times \textit{Hematology Indices}$. The second tensor dataset contains the MRI data: $\eX_2\in \mathbb{R}^{19\times 4 \times 14}$, organized as $\textit{Monkey Infants} \times \textit{DTI Parameters} \times \textit{Brain Regions}$. While the MRI data were complete, the hematology tensor $\eX_1$ exhibited missing entries at certain time points that we would like to impute. We aim to apply the MULTIFAC method to identify whether there is common structures between $\eX_1$ and $\eX_2$, which could potentially indicate associations between hematological profiles and difference in brain function after the onset of iron deficiency.   

\subsection{Results}
\begin{figure}[H]
    \centering
    \includegraphics[width=\linewidth]{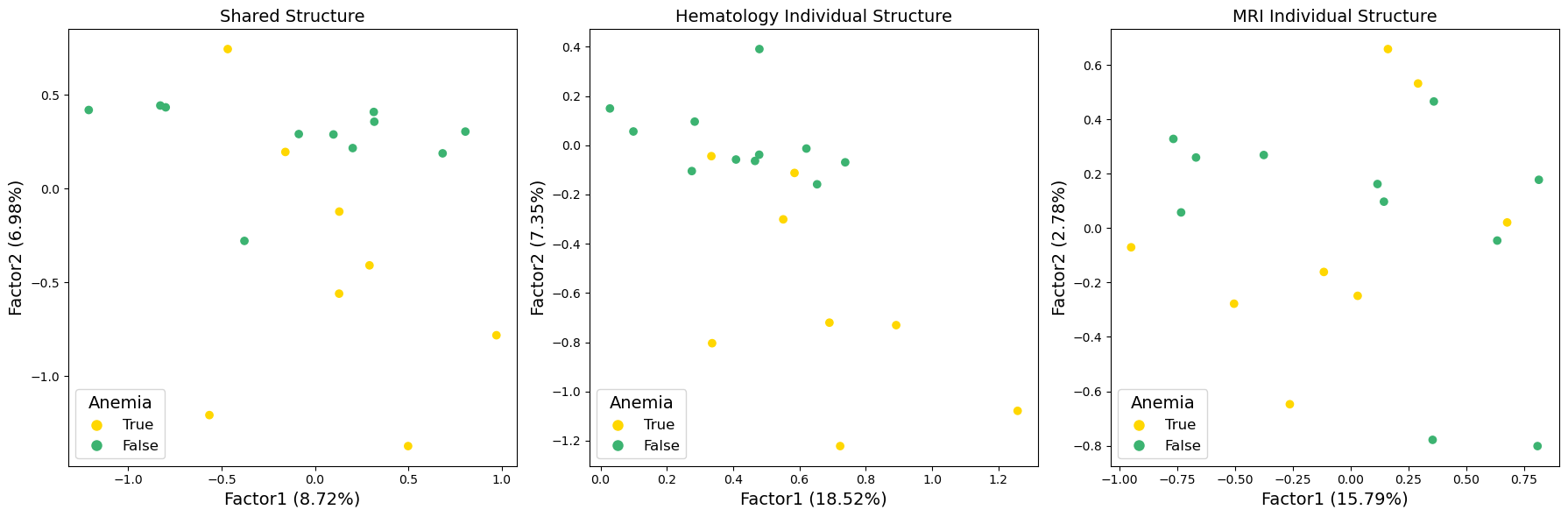}    
    \caption{Top Sample Loadings in Different Structures}
    \label{fig:2.1}
\end{figure}

We summarized the proportion of variance explained by various structures and their estimated tensor ranks in Table \ref{table:6_1}. For the analysis based on the common ID subset, 20–40\% of the variability was captured by the shared structures between hematology and MRI data. For each structure, we visualized the top two sample loadings associated with the rank-1 components explaining the largest variance in Figure \ref{fig:2.1}. The anemia status of each sample was indicated by color. The shared structure plots exhibited some patterns associated with anemia development but did not provide clear discrimination. Notably, the individual hematology loadings clearly discriminated anemia status, whereas the individual MRI loadings did not. This is expected, as hematology is a more directly related to anemia status (so any patterns associated with anemia in MRI would also be shared with hematology).     

\begin{figure}[H]
    \centering
    \includegraphics[width=0.95\linewidth]{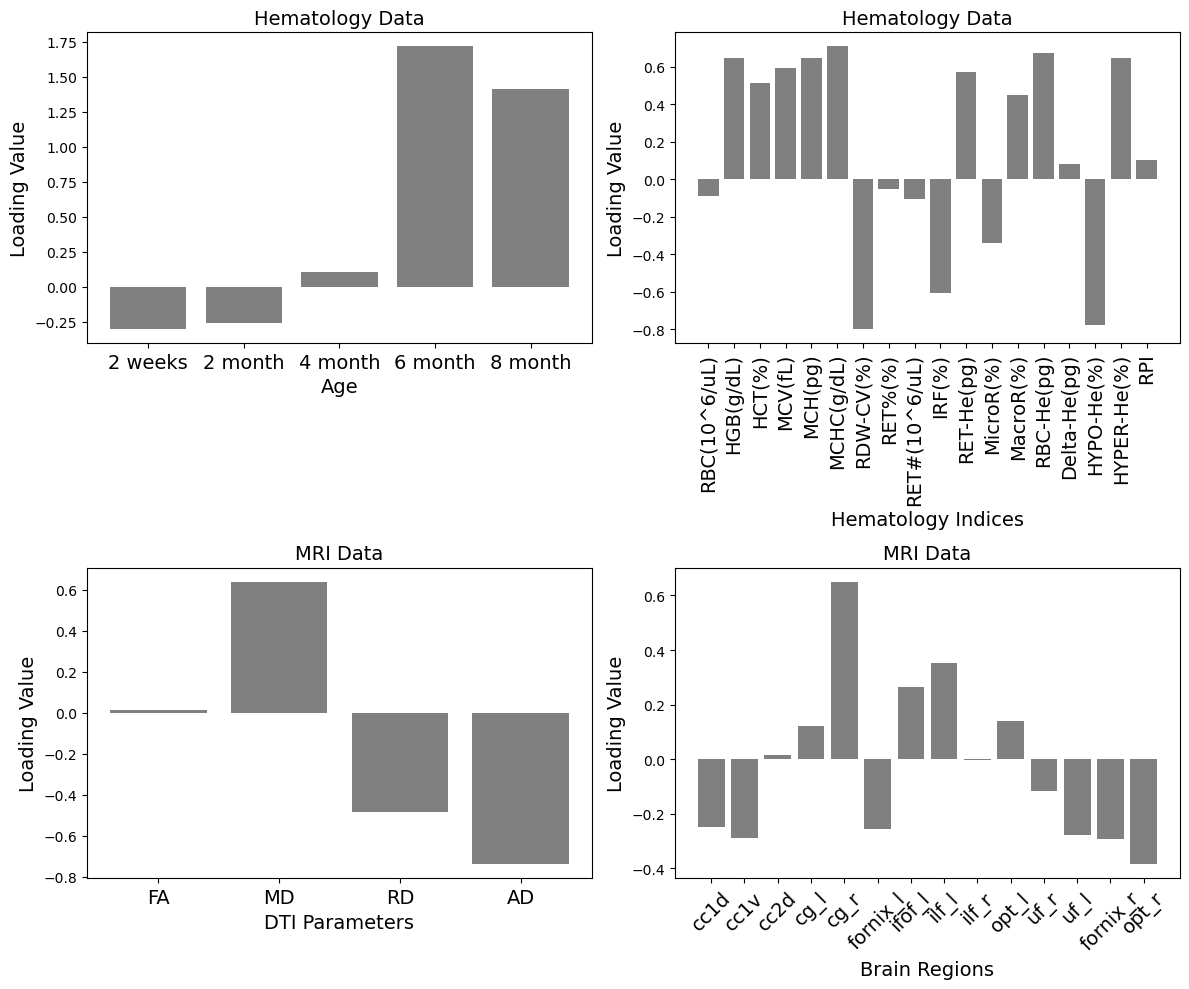}    
    \caption{Loading plots for the second component of in shared structure}
    \label{fig:2.2}
\end{figure}

\begin{figure}[H]
    \centering
    \includegraphics[width=0.95\linewidth]{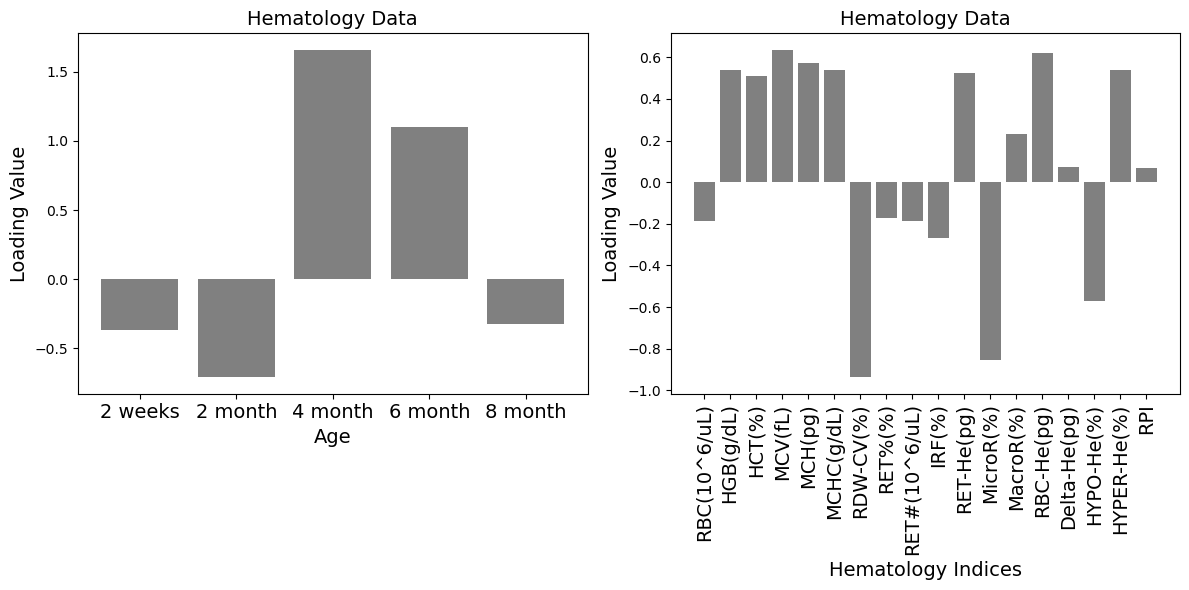}    
    \caption{Loading plots for the second component of the individual structure for hematology.}
    \label{fig:2.3}
\end{figure}

Interestingly, the sample loadings for the second rank-1 component demonstrated better discrimination of anemia status than those for the first component for both the shared and individual structure. Figures \ref{fig:2.2} and \ref{fig:2.3} present the corresponding loadings for other dimensions. For the second shared component, the 6 and 8 month time-points have the largest loadings for the hematology data; these time-points are closer to the MRI data collection, which occurred at approximately 12-months. The second component for the hematology-specific structure showed the best discrimination  by anemia status, and had the largest loadings at 4 and 6 months; these time points are generally when the effects of anemia are greatest, prior to recovery.

\begin{table}[!h]
\caption{Application of MULTIFAC to Sysmex Hematology and MRI data. Each cell reports the proportion of variance explained by different structures, with the estimated rank shown inside parentheses.}
\centering
\begin{tabular}{|l|l|l|l|}
\hline
Dataset, Structure & Total & Shared & Individual\\ \hline
Hematology ($\eX_1$) & 0.757 (13) & 0.244 (6) & 0.451 (7) \\ \hline
MRI ($\eX_2$) & 0.708 (16) & 0.373 (6) & 0.311 (10) \\ \hline
\end{tabular}
\label{table:6_1}
\end{table}

\section{Discussion}
This work introduces a novel framework for tensor decomposition, designed specifically to analyze multiple tensors linked through their first mode. The proposed method effectively distinguishes shared and individual structures across datasets while handling missing data scenarios, such as entry-wise and tensor-wise missing patterns. Through rigorous simulations, the method demonstrated superior performance over existing approaches, consistently achieving lower RSE values across a wide range of conditions. Notably, the penalty framework employed enables rank sparsity, facilitating automatic determination of tensor ranks without compromising model robustness.

While the proposed method offers significant improvements, there are areas requiring further exploration. First, the current grid search approach for selecting penalty factors is computationally intensive, particularly for large datasets. Future efforts could focus on incorporating advanced optimization techniques to expedite the penalty tuning process while maintaining robustness. Additionally, empirical methods for penalty selection could be replaced with theoretically grounded approaches, such as variational Bayesian models that directly treat the penalty as a prior parameter or random tensor theory to derive penalties systematically.  Moreover, the current framework is tailored for tensors linked along their first mode. Extending this model to scenarios where subsets of tensors share multiple modes presents an exciting challenge. Such an extension would enhance the applicability of the method to more complex multi-way data structures but would necessitate the development of more efficient optimization algorithms to manage increased computational complexity.

\section*{Supplementary materials} 
Python code to perform MULTIFAC with examples are available at \url{https://github.com/zhiyu-kang/MULTIFAC}. 

\section*{Acknowledgment}
This work was supported by the NIH National Institute of General Medical Sciences
(NIGMS) grant R01-GM130622. The data used in this study  was funded by grants from the National Institute of Health/Eunice Kennedy Shriver National Institute of Child Health and Development (HD089989) and Sysmex America, Inc., Lincolnshire, Illinois. The funding agencies were not involved in study design, execution, data collection, interpretation and dissemination.

\appendix
\section{Appendix}
\label{appendix:A}
\subsection{Proof of Theorem \ref{thm:1}}
\begin{proof}
Consider the simplest problem where we penalize a rank-1 CP decomposition for a three-way tensor $\eX: m\times n \times k$. Denote the factor matrices as $\ba\in \mathbb{R}^{m}$, $\bb\in \mathbb{R}^{n}$ and $\bc\in \mathbb{R}^{k}$. 
By the relationship between arithmetic mean and geometric mean, we have:
\begin{equation}\label{eq:10}
    \big\|\ba\big\|_F^2 + \big\|\bb\big\|_F^2 + \big\|\bc\big\|_F^2 \geq 3 \left(\big\|\ba\big\|_F  \big\|\bb\big\|_F  \big\|\bc\big\|_F\right)^{2/3}
\end{equation}
Equity is obtained if and only if $\|\ba\|_F = \|\bb\|_F = \|\bc\|_F$.
We claim that: 
\begin{equation}\label{eq:11}
\begin{aligned}
    &\text{min}\big\|\eX-[\![\ba, \bb,\bc]\!] \big\|_F^2  +\sigma\left(\big\|\ba\big\|_F^2 + \big\|\bb\big\|_F^2 + \big\|\bc\big\|_F^2\right)\\
    = \operatorname*{min}&_{\|\ba\|_F = \|\bb\|_F = \|\bc\|_F} \big\|\eX-[\![\ba, \bb,\bc ]\!] \big\|_F^2  + 3\sigma \left(\big\|\ba\big\|_F  \big\|\bb\big\|_F  \big\|\bc\big\|_F\right)^{2/3}
\end{aligned}
\end{equation}
If the solution to the left-hand side of \ref{eq:11} does not satisfy the condition: $\|\ba\|_F = \|\bb\|_F = \|\bc\|_F$, we can rescale the factor matrices such that the condition holds but the Frobenius norm of residual stays unchanged. Thus, this solution is not the argument of minimal according to \ref{eq:10}. By contradiction, the condition must be satisfied and equation \ref{eq:11} will hold.
Hence, the problem is equivalent to the constraint problem:
\begin{equation*}
\begin{aligned}
    \operatorname*{min}&_{\|\ba\|_F = \|\bb\|_F = \|\bc\|_F} \big\|\eX-[\![\ba, \bb,\bc ]\!] \big\|_F^2  +3\sigma \left(\big\|\ba\big\|_F  \big\|\bb\big\|_F  \big\|\bc\big\|_F\right)^{2/3}\\
    &=\text{min} \big\|\eX-\lambda \tilde\ba \circ \tilde\bb \circ \tilde\bc \big\|_F^2  +3\sigma \lambda^{2/3}, 
\end{aligned}
\end{equation*}
where $\tilde\ba={\ba}/{\|\ba\|_F}$, $\tilde\bb={\bb}/{\|\bb\|_F}$ and $\tilde\bc={\bc}/{\|\bc\|_F}$. The $L_{2/3}$ penalty on the factor $\lambda$ may shrink this parameter estimate to 0. 

For a rank $R$ CP decomposition, the result is similar:
\begin{equation*}
\begin{aligned}
    &\text{min}\big\|\eX-[\![\cA, \cB,\cC ]\!] \big\|_F^2  +\sigma\left(\big\|\cA\big\|_F^2 + \big\|\cB\big\|_F^2 + \big\|\cC\big\|_F^2\right)\\
    &=\operatorname*{min}_{\| \ba_i\|_F = \| \bb_i\|_F = \| \bc_i\|_F} \big\|\eX-[\![\cA, \cB,\cC ]\!] \big\|_F^2  +3\sigma \sum_{i=1}^R \left(\big\| \ba_i\big\|_F  \big\| \bb_i\big\|_F  \big\| \bc_i\big\|_F\right)^{2/3}\\
    &=\text{min} \big\|\eX-\sum_{i=1}^R\lambda_i  \tilde\ba_i \circ \tilde\bb_i \circ \tilde\bc_i \big\|_F^2  +3\sigma \sum_{i=1}^R\lambda_i^{2/3},\\
\end{aligned}
\end{equation*}
where $\ba_i$, $\bb_i$ and $\bc_i$ are the ith columns of factor matrices $\cA$, $\cB$ and $\cC$ respectively. We rescale the factor matrices such that each column has unit Frobenius norm. $\tilde\ba_i$, $\tilde\bb_i$ and $\tilde\bc_i$ are the $i$-th columns of the rescaled factor matrices. Same as in the rank-1 case, the $L_{2/3}$ penalty yields rank sparsity. This conclusion can be easily generalized to multi-way tensor factorization. An equivalent form of the $L_2$ penalized problem can be written as a $L_{2/n}$ penalized problem:
$$
\text{min} \big\|\eX-[\![\bm \lambda; \tilde\cA_{1},\ldots,\tilde\cA_n ]\!] \big\|_F^2  +N\sigma \|\bm \lambda\|_{2/N}^{2/N}.
$$
Since $2/n<1$, the penalty term on the factor vector $\bm \lambda$ yields sparsity in $\bm \lambda$, thus it lead to rank sparsity of the tensor.
\end{proof}

\subsection{Proof of Theorem \ref{thm:2}}
\begin{proof}
In order to prove the solutions of factor matrices are the same for two minimization problems, we first simplify the penalized problem by ignoring terms that are unrelated to $(\lambda, \ba_1, \ba_2, \ldots, \ba_N)$:
\begin{equation*}
    \begin{aligned}
        &\text{min }\big \| \eX - [\![ \lambda; \ba_1, \ba_2, \ldots, \ba_N]\!] \big \|_F^2 + N\sigma \|\lambda\|_{2/N}^{2/N}\\
        =\ & \text{min } tr[(\eX_{(1)}-\lambda \ba_1(\ba_N\odot \ba_{N-1}\odot \cdots \odot \ba_2)^\top)^\top(\eX_{(1)}-\lambda \ba_1(\ba_N\odot \ba_{N-1}\odot \cdots \odot \ba_2)^\top)] + N\sigma \lambda^{2/N}\\
        =\ & \text{min } \lambda^2 tr((\ba_N\odot  \cdots \odot \ba_2)\ba_1^\top \ba_1(\ba_N\odot  \cdots \odot \ba_2)^\top)- 2\lambda tr(\eX_{(1)}^\top \ba_1(\ba_N\odot  \cdots \odot \ba_2)^\top)
        + N\sigma \lambda^{2/N}\\
        =\ & \text{min }\lambda^2 tr((\ba_N\odot  \cdots \odot \ba_2)^\top(\ba_N\odot  \cdots \odot \ba_2)\ba_1^\top \ba_1) - 2\lambda tr(\eX_{(1)}^\top \ba_1(\ba_N\odot  \cdots \odot \ba_2)^\top) + N\sigma \lambda^{2/N}\\
        =\ & \text{min }\lambda^2-2\lambda tr(\eX_{(1)}^\top \ba_1(\ba_N\odot  \cdots \odot \ba_2)^\top)+N\sigma \lambda^{2/N}
    \end{aligned}
\end{equation*}
Note that the unpenalized problem can be simplified in the same way as:
\begin{equation*}
    \begin{aligned}
        & \text{min }\big \| \eX - [\![ \lambda; \ba_1, \ba_2, \ldots, \ba_N]\!] \big \|_F^2\\
        =\ & \text{min }tr[(\eX_{(1)}-\lambda \ba_1(\ba_N\odot  \cdots \odot \ba_2)^\top)^\top(\eX_{(1)}-\lambda \ba_1(\ba_N\odot  \cdots \odot \ba_2)^\top)]\\
        =\ & \text{min }\lambda^2 -2\lambda tr(\eX_{(1)}^\top \ba_1(\ba_N\odot  \cdots \odot \ba_2)^\top) 
    \end{aligned}
\end{equation*}
Since both problems have only one term related to $(\ba_1, \ba_2, \ldots, \ba_N)$: $-2\lambda tr(\eX_{(1)}^\top \ba_1(\ba_N\odot  \cdots \odot \ba_2)^\top) $ and $\lambda \geq 0$, their solution to $(\ba_1, \ba_2, \ldots, \ba_N)$ should be the same. When given the solution $(\hat{\ba}_1, \hat{\ba}_2, \ldots, \hat{\ba}_N)$,
$$
\operatorname*{min}_{\lambda}\big \| \eX - [\![ \lambda; \hat{\ba}_1, \hat{\ba}_2, \ldots, \hat{\ba}_N]\!] \big \|_F^2 =\operatorname*{min}_{\lambda} (\lambda-tr(\eX_{(1)}^\top \hat{\ba}_1(\hat{\ba}_N\odot  \cdots \odot \hat{\ba}_2)^\top))^2 - tr(\eX_{(1)}^\top \hat{\ba}_1(\hat{\ba}_N\odot  \cdots \odot \hat{\ba}_2)^\top) ^2,
$$
meaning that the objective function is minimized at $tr(\eX_{(1)}^\top \hat{\ba}_1(\hat{\ba}_N\odot  \cdots \odot \hat{\ba}_2)^\top)= \hat{\lambda}$.
For the penalized problem we can have 
$$
\hat{\lambda}_p = \operatorname*{argmin}_{\lambda} \big \| \eX - [\![ \lambda; \hat{\ba}_1, \hat{\ba}_2, \ldots, \hat{\ba}_N]\!] \big \|_F^2 + N\sigma \|\lambda\|_{2/N}^{2/N} = \operatorname*{argmin}_{\lambda} (\lambda - \hat{\lambda})^2 + N\sigma \lambda^{2/N}.
$$
\end{proof}

\subsection{Proof of Theorem \ref{thm:3}}
\begin{proof}
Denote the $r$-th column of shared factor matrix $\cA_0$ as $\ba_{0r}$ and the the $r$-th column of remaining factor matrices $\cA_{i}^{(k)}$, as $\ba_{ir}^{(k)}$, for $i=1,\ldots, N_k$. Similar to the idea in Theorem 1, we can have 
\begin{align*}
    &\big\|\cA_0\big\|_F^2 + \sum_{k=1}^K\sum_{i=1}^{N_k} \big\|\cA_i^{(k)}\big\|_F^2\\
    = \ & \sum_{r=1}^R \|\ba_{0r}\|_F^2 + \sum_{k=1}^K\sum_{r=1}^R\sum_{i=1}^{N_k} \| \ba_{ir}^{(k)}\|_F^2\\
    \geq\  & \sum_{r=1}^R \|\ba_{0r}\|_F^2 + \sum_{k=1}^K\sum_{r=1}^R N_k \big(\prod_{k=1}^{N_k}\|\ba_{ir}^{(k)}\|_F\big)^{2/N_k}\\
    =\ & \|\bm \lambda_0\|_2^2 + \sum_{k=1}^K N_k\|\bm\lambda_{(0)i} \|_{2/N_k}^{2/N_k}
\end{align*}

Equality holds if and only if the condition $\bm \ba_{ir}^{(1)} = \ba_{ir}^{(2)} = \cdots =\ba_{ir}^{(K)}$ is satisfied for $i=1,\ldots,K$ and $r=1,\ldots,R$. These conditions can be easily satisfied by simply scaling columns of the factor matrices $\ba_{ir}^{(k)}$.
\end{proof}



\begin{thebibliography}{24}
\ifx \bisbn   \undefined \def \bisbn  #1{ISBN #1}\fi
\ifx \binits  \undefined \def \binits#1{#1}\fi
\ifx \bauthor  \undefined \def \bauthor#1{#1}\fi
\ifx \batitle  \undefined \def \batitle#1{#1}\fi
\ifx \bjtitle  \undefined \def \bjtitle#1{#1}\fi
\ifx \bvolume  \undefined \def \bvolume#1{\textbf{#1}}\fi
\ifx \byear  \undefined \def \byear#1{#1}\fi
\ifx \bissue  \undefined \def \bissue#1{#1}\fi
\ifx \bfpage  \undefined \def \bfpage#1{#1}\fi
\ifx \blpage  \undefined \def \blpage #1{#1}\fi
\ifx \burl  \undefined \def \burl#1{\textsf{#1}}\fi
\ifx \doiurl  \undefined \def \doiurl#1{\url{https://doi.org/#1}}\fi
\ifx \betal  \undefined \def \betal{\textit{et al.}}\fi
\ifx \binstitute  \undefined \def \binstitute#1{#1}\fi
\ifx \binstitutionaled  \undefined \def \binstitutionaled#1{#1}\fi
\ifx \bctitle  \undefined \def \bctitle#1{#1}\fi
\ifx \beditor  \undefined \def \beditor#1{#1}\fi
\ifx \bpublisher  \undefined \def \bpublisher#1{#1}\fi
\ifx \bbtitle  \undefined \def \bbtitle#1{#1}\fi
\ifx \bedition  \undefined \def \bedition#1{#1}\fi
\ifx \bseriesno  \undefined \def \bseriesno#1{#1}\fi
\ifx \blocation  \undefined \def \blocation#1{#1}\fi
\ifx \bsertitle  \undefined \def \bsertitle#1{#1}\fi
\ifx \bsnm \undefined \def \bsnm#1{#1}\fi
\ifx \bsuffix \undefined \def \bsuffix#1{#1}\fi
\ifx \bparticle \undefined \def \bparticle#1{#1}\fi
\ifx \barticle \undefined \def \barticle#1{#1}\fi
\bibcommenthead
\ifx \bconfdate \undefined \def \bconfdate #1{#1}\fi
\ifx \botherref \undefined \def \botherref #1{#1}\fi
\ifx \url \undefined \def \url#1{\textsf{#1}}\fi
\ifx \bchapter \undefined \def \bchapter#1{#1}\fi
\ifx \bbook \undefined \def \bbook#1{#1}\fi
\ifx \bcomment \undefined \def \bcomment#1{#1}\fi
\ifx \oauthor \undefined \def \oauthor#1{#1}\fi
\ifx \citeauthoryear \undefined \def \citeauthoryear#1{#1}\fi
\ifx \endbibitem  \undefined \def \endbibitem {}\fi
\ifx \bconflocation  \undefined \def \bconflocation#1{#1}\fi
\ifx \arxivurl  \undefined \def \arxivurl#1{\textsf{#1}}\fi
\csname PreBibitemsHook\endcsname

\bibitem[\protect\citeauthoryear{Acar et~al.}{2011}]{acar2011scalable}
\begin{barticle}
\bauthor{\bsnm{Acar}, \binits{E.}},
\bauthor{\bsnm{Dunlavy}, \binits{D.M.}},
\bauthor{\bsnm{Kolda}, \binits{T.G.}},
\bauthor{\bsnm{M{\o}rup}, \binits{M.}}:
\batitle{Scalable tensor factorizations for incomplete data}.
\bjtitle{Chemometrics and Intelligent Laboratory Systems}
\bvolume{106}(\bissue{1}),
\bfpage{41}--\blpage{56}
(\byear{2011})
\end{barticle}
\endbibitem

\bibitem[\protect\citeauthoryear{Acar et~al.}{2011}]{acar_2011}
\begin{botherref}
\oauthor{\bsnm{Acar}, \binits{E.}},
\oauthor{\bsnm{Kolda}, \binits{T.G.}},
\oauthor{\bsnm{Dunlavy}, \binits{D.M.}}:
All-at-once optimization for coupled matrix and tensor factorizations.
ArXiv
\textbf{abs/1105.3422}
(2011)
\end{botherref}
\endbibitem

\bibitem[\protect\citeauthoryear{Acar et~al.}{2013}]{acar_2013}
\begin{botherref}
\oauthor{\bsnm{Acar}, \binits{E.}},
\oauthor{\bsnm{Lawaetz}, \binits{A.J.}},
\oauthor{\bsnm{Rasmussen}, \binits{M.A.}},
\oauthor{\bsnm{Bro}, \binits{R.}}:
Structure-revealing data fusion model with applications in metabolomics.
2013 35th Annual International Conference of the IEEE Engineering in Medicine
  and Biology Society (EMBC),
6023--6026
(2013)
\end{botherref}
\endbibitem

\bibitem[\protect\citeauthoryear{Argelaguet et~al.}{2018}]{argelaguet2018multi}
\begin{barticle}
\bauthor{\bsnm{Argelaguet}, \binits{R.}},
\bauthor{\bsnm{Velten}, \binits{B.}},
\bauthor{\bsnm{Arnol}, \binits{D.}},
\bauthor{\bsnm{Dietrich}, \binits{S.}},
\bauthor{\bsnm{Zenz}, \binits{T.}},
\bauthor{\bsnm{Marioni}, \binits{J.C.}},
\bauthor{\bsnm{Buettner}, \binits{F.}},
\bauthor{\bsnm{Huber}, \binits{W.}},
\bauthor{\bsnm{Stegle}, \binits{O.}}:
\batitle{Multi-omics factor analysis—a framework for unsupervised integration
  of multi-omics data sets}.
\bjtitle{Molecular systems biology}
\bvolume{14}(\bissue{6}),
\bfpage{8124}
(\byear{2018})
\end{barticle}
\endbibitem

\bibitem[\protect\citeauthoryear{Breiman et~al.}{1984}]{1se_rule}
\begin{botherref}
\oauthor{\bsnm{Breiman}, \binits{L.}},
\oauthor{\bsnm{Friedman}, \binits{J.}},
\oauthor{\bsnm{Stone}, \binits{C.J.}},
\oauthor{\bsnm{Olshen}, \binits{R.A.}}:
Classification and Regression Trees.
Taylor \& Francis
(1984)
\end{botherref}
\endbibitem

\bibitem[\protect\citeauthoryear{Carroll and Chang}{1970}]{carroll_chang_1970}
\begin{barticle}
\bauthor{\bsnm{Carroll}, \binits{J.D.}},
\bauthor{\bsnm{Chang}, \binits{J.-J.}}:
\batitle{Analysis of individual differences in multidimensional scaling via an
  n-way generalization of “eckart-young” decomposition}.
\bjtitle{Psychometrika}
\bvolume{35}(\bissue{3}),
\bfpage{283}--\blpage{319}
(\byear{1970})
\end{barticle}
\endbibitem

\bibitem[\protect\citeauthoryear{De~Vito et~al.}{2021}]{de2021bayesian}
\begin{barticle}
\bauthor{\bsnm{De~Vito}, \binits{R.}},
\bauthor{\bsnm{Bellio}, \binits{R.}},
\bauthor{\bsnm{Trippa}, \binits{L.}},
\bauthor{\bsnm{Parmigiani}, \binits{G.}}:
\batitle{Bayesian multistudy factor analysis for high-throughput biological
  data}.
\bjtitle{The annals of applied statistics}
\bvolume{15}(\bissue{4}),
\bfpage{1723}--\blpage{1741}
(\byear{2021})
\end{barticle}
\endbibitem

\bibitem[\protect\citeauthoryear{Fu}{1998}]{fu_1998}
\begin{barticle}
\bauthor{\bsnm{Fu}, \binits{W.J.}}:
\batitle{Penalized regressions: The bridge versus the lasso}.
\bjtitle{Journal of Computational and Graphical Statistics}
\bvolume{7}(\bissue{3}),
\bfpage{397}--\blpage{416}
(\byear{1998})
\end{barticle}
\endbibitem

\bibitem[\protect\citeauthoryear{Gaynanova and
  Li}{2019}]{gaynanova_li_slide_2019}
\begin{barticle}
\bauthor{\bsnm{Gaynanova}, \binits{I.}},
\bauthor{\bsnm{Li}, \binits{G.}}:
\batitle{Structural learning and integrative decomposition of multi-view data}.
\bjtitle{Biometrics}
\bvolume{75}(\bissue{4}),
\bfpage{1121}--\blpage{1132}
(\byear{2019})
\end{barticle}
\endbibitem

\bibitem[\protect\citeauthoryear{Harshman}{1970}]{harshman_1970}
\begin{barticle}
\bauthor{\bsnm{Harshman}, \binits{R.A.}}:
\batitle{Foundations of the parafac procedure: Models and conditions for an
  “explanatory” multimodal factor analysis}.
\bjtitle{UCLA Working Papers in Phonetics}
\bvolume{16},
\bfpage{1}--\blpage{84}
(\byear{1970}).
\bcomment{University Microfilms, Ann Arbor, Michigan, No. 10,085}
\end{barticle}
\endbibitem

\bibitem[\protect\citeauthoryear{Hitchcock}{1927}]{hitchcock_1927}
\begin{barticle}
\bauthor{\bsnm{Hitchcock}, \binits{F.L.}}:
\batitle{The expression of a tensor or a polyadic as a sum of products}.
\bjtitle{Journal of Mathematics and Physics}
\bvolume{6}(\bissue{1-4}),
\bfpage{164}--\blpage{189}
(\byear{1927})
\doiurl{10.1002/sapm192761164}
{\href{https://arxiv.org/abs/https://onlinelibrary.wiley.com/doi/pdf/10.1002/sapm192761164}{{https://onlinelibrary.wiley.com/doi/pdf/10.1002/sapm192761164}}}
\end{barticle}
\endbibitem

\bibitem[\protect\citeauthoryear{Hoerl and
  Kennard}{1970}]{hoerl_kennard_1970_ridge}
\begin{barticle}
\bauthor{\bsnm{Hoerl}, \binits{A.E.}},
\bauthor{\bsnm{Kennard}, \binits{R.W.}}:
\batitle{Ridge regression: Biased estimation for nonorthogonal problems}.
\bjtitle{Technometrics}
\bvolume{12}(\bissue{1}),
\bfpage{55}--\blpage{67}
(\byear{1970})
\end{barticle}
\endbibitem

\bibitem[\protect\citeauthoryear{Kolda and
  Bader}{2009}]{kolda_tensor_review_2009}
\begin{barticle}
\bauthor{\bsnm{Kolda}, \binits{T.G.}},
\bauthor{\bsnm{Bader}, \binits{B.W.}}:
\batitle{Tensor decompositions and applications}.
\bjtitle{SIAM Review}
\bvolume{51}(\bissue{3}),
\bfpage{455}--\blpage{500}
(\byear{2009})
\end{barticle}
\endbibitem

\bibitem[\protect\citeauthoryear{Khan et~al.}{2014}]{khan_bayesian_2014}
\begin{barticle}
\bauthor{\bsnm{Khan}, \binits{S.A.}},
\bauthor{\bsnm{Lepp{\"a}aho}, \binits{E.}},
\bauthor{\bsnm{Kaski}, \binits{S.}}:
\batitle{Bayesian multi-tensor factorization}.
\bjtitle{Machine Learning}
\bvolume{105},
\bfpage{233}--\blpage{253}
(\byear{2014})
\end{barticle}
\endbibitem

\bibitem[\protect\citeauthoryear{Kruskal}{1977}]{kruskal_1977}
\begin{barticle}
\bauthor{\bsnm{Kruskal}, \binits{J.B.}}:
\batitle{Three-way arrays: rank and uniqueness of trilinear decompositions,
  with application to arithmetic complexity and statistics}.
\bjtitle{Linear Algebra and its Applications}
\bvolume{18}(\bissue{2}),
\bfpage{95}--\blpage{138}
(\byear{1977})
\doiurl{10.1016/0024-3795(77)90069-6}
\end{barticle}
\endbibitem

\bibitem[\protect\citeauthoryear{Lock et~al.}{2013}]{lock_jive_2013}
\begin{barticle}
\bauthor{\bsnm{Lock}, \binits{E.F.}},
\bauthor{\bsnm{Hoadley}, \binits{K.A.}},
\bauthor{\bsnm{Marron}, \binits{J.}},
\bauthor{\bsnm{Nobel}, \binits{A.B.}}:
\batitle{Joint and individual variance explained ({JIVE}) for integrated
  analysis of multiple data types}.
\bjtitle{Annals of Applied Statistics}
\bvolume{7}(\bissue{1}),
\bfpage{523}--\blpage{542}
(\byear{2013})
\end{barticle}
\endbibitem

\bibitem[\protect\citeauthoryear{Lock et~al.}{2022}]{lock_park_2020}
\begin{barticle}
\bauthor{\bsnm{Lock}, \binits{E.F.}},
\bauthor{\bsnm{Park}, \binits{J.Y.}},
\bauthor{\bsnm{Hoadley}, \binits{K.A.}}:
\batitle{Bidimensional linked matrix factorization for pan-omics pan-cancer
  analysis}.
\bjtitle{The Annals of Applied Statistics}
\bvolume{16 1},
\bfpage{193}--\blpage{215}
(\byear{2022})
\end{barticle}
\endbibitem

\bibitem[\protect\citeauthoryear{Mazumder et~al.}{2010}]{mazumder_2010}
\begin{barticle}
\bauthor{\bsnm{Mazumder}, \binits{R.}},
\bauthor{\bsnm{Hastie}, \binits{T.}},
\bauthor{\bsnm{Tibshirani}, \binits{R.}}:
\batitle{Spectral regularization algorithms for learning large incomplete
  matrices}.
\bjtitle{Journal of Machine Learning Research}
\bvolume{11}(\bissue{80}),
\bfpage{2287}--\blpage{2322}
(\byear{2010})
\end{barticle}
\endbibitem

\bibitem[\protect\citeauthoryear{Park and Lock}{2020}]{park_lock_2019}
\begin{barticle}
\bauthor{\bsnm{Park}, \binits{J.Y.}},
\bauthor{\bsnm{Lock}, \binits{E.F.}}:
\batitle{{Integrative Factorization of Bidimensionally Linked Matrices}}.
\bjtitle{Biometrics}
\bvolume{76}(\bissue{1}),
\bfpage{61}--\blpage{74}
(\byear{2020})
\end{barticle}
\endbibitem

\bibitem[\protect\citeauthoryear{Palzer et~al.}{2022}]{sjive_2022}
\begin{barticle}
\bauthor{\bsnm{Palzer}, \binits{E.F.}},
\bauthor{\bsnm{Wendt}, \binits{C.H.}},
\bauthor{\bsnm{Bowler}, \binits{R.P.}},
\bauthor{\bsnm{Hersh}, \binits{C.P.}},
\bauthor{\bsnm{Safo}, \binits{S.E.}},
\bauthor{\bsnm{Lock}, \binits{E.F.}}:
\batitle{sjive: Supervised joint and individual variation explained}.
\bjtitle{Computational Statistics \& Data Analysis}
\bvolume{175},
\bfpage{107547}
(\byear{2022})
\end{barticle}
\endbibitem

\bibitem[\protect\citeauthoryear{Rao et~al.}{2023}]{rao_lock_etal_2023}
\begin{barticle}
\bauthor{\bsnm{Rao}, \binits{R.B.}},
\bauthor{\bsnm{Lubach}, \binits{G.R.}},
\bauthor{\bsnm{Ennis-Czerniak}, \binits{K.M.}},
\bauthor{\bsnm{Lock}, \binits{E.F.}},
\bauthor{\bsnm{Kling}, \binits{P.J.}},
\bauthor{\bsnm{Georgieff}, \binits{M.K.}},
\bauthor{\bsnm{Coe}, \binits{C.L.}}:
\batitle{Reticulocyte hemoglobin equivalent has comparable predictive accuracy
  as conventional serum iron indices for predicting iron deficiency and anemia
  in a nonhuman primate model of infantile iron deficiency}.
\bjtitle{The Journal of Nutrition}
\bvolume{153}(\bissue{1}),
\bfpage{148}--\blpage{157}
(\byear{2023})
\end{barticle}
\endbibitem

\bibitem[\protect\citeauthoryear{Sorber et~al.}{2015}]{sorber_sdf_2015}
\begin{barticle}
\bauthor{\bsnm{Sorber}, \binits{L.}},
\bauthor{\bsnm{Van~Barel}, \binits{M.}},
\bauthor{\bsnm{De~Lathauwer}, \binits{L.}}:
\batitle{Structured data fusion}.
\bjtitle{IEEE Journal of Selected Topics in Signal Processing}
\bvolume{9}(\bissue{4}),
\bfpage{586}--\blpage{600}
(\byear{2015})
\end{barticle}
\endbibitem

\bibitem[\protect\citeauthoryear{Tibshirani}{1996}]{tibshirani_1996_lasso}
\begin{barticle}
\bauthor{\bsnm{Tibshirani}, \binits{R.}}:
\batitle{Regression shrinkage and selection via the lasso}.
\bjtitle{Journal of the Royal Statistical Society: Series B (Methodological)}
\bvolume{58}(\bissue{1}),
\bfpage{267}--\blpage{288}
(\byear{1996})
\end{barticle}
\endbibitem

\bibitem[\protect\citeauthoryear{Wang and Lock}{2024}]{wang_lock_2024}
\begin{barticle}
\bauthor{\bsnm{Wang}, \binits{J.}},
\bauthor{\bsnm{Lock}, \binits{E.F.}}:
\batitle{{Multiple augmented reduced rank regression for pan-cancer analysis}}.
\bjtitle{Biometrics}
\bvolume{80}(\bissue{1}),
\bfpage{002}
(\byear{2024})
\end{barticle}
\endbibitem

\end{thebibliography}
\end{document}